\pdfoutput=1

\documentclass[runningheads]{llncs}
\usepackage{graphicx}

\usepackage{tikz}
\usepackage{comment}
\usepackage{amsmath,amssymb} %
\usepackage{color}

\usepackage{booktabs}
\usepackage{cite}
\usepackage{multirow}
\usepackage{orcidlink}

\usepackage[accsupp]{axessibility}  %

\newcommand{\PAR}[1]{\vskip2pt \noindent{\bf #1~}}

\usepackage{xspace}

\newcommand{\ie}{\textit{i.e.}\xspace}
\newcommand{\eg}{\textit{e.g.}\xspace}

\newcommand{\Eg}{\textit{E.g.}\xspace}

\newcommand{\cf}{\textit{cf.}\xspace}
\newcommand{\etc}{\textit{etc.}\xspace}
\newcommand{\wrt}{\textit{w.r.t.}\xspace}

\begin{document}
\pagestyle{headings}
\mainmatter
\def\ECCVSubNumber{5648}  %

\title{MeshLoc: Mesh-Based Visual Localization} %

\titlerunning{MeshLoc: Mesh-Based Visual Localization}
\author{Vojtech Panek\inst{1,2}\orcidlink{0000-0003-0601-7682} \and
Zuzana Kukelova\inst{3}\orcidlink{0000-0002-1916-8829} \and
Torsten Sattler\inst{2}\orcidlink{0000-0001-9760-4553}}
\authorrunning{V. Panek et al.}
\institute{Faculty of Electrical Engineering, Czech Technical University (CTU) in Prague \and Czech Institute of Informatics, Robotics and Cybernetics, CTU in Prague \and Visual Recognition Group, Faculty of Electrical Engineering, CTU in Prague}
\maketitle

\begin{abstract}
Visual localization, \ie, the problem of camera pose estimation, is a central component of applications such as autonomous robots and augmented reality systems. A dominant approach in the literature, shown to scale to large scenes and to handle complex illumination and seasonal changes, is based on local features extracted from images. The scene representation is a sparse Structure-from-Motion point cloud that is tied to a specific local feature. Switching to another feature type requires an expensive feature matching step between the database images used to construct the point cloud. In this work, we thus explore a more flexible alternative based on dense 3D meshes that does not require features matching between database images to build the scene representation. We show that this approach can achieve state-of-the-art results. We further show that surprisingly competitive results can be obtained when extracting features on renderings of these meshes, without any neural rendering stage, and even when rendering raw scene geometry without color or texture. Our results show that dense 3D model-based representations are a promising alternative to existing representations and point to interesting and challenging directions for future research. 

\end{abstract}

\section{Introduction}
\label{sec:intro}
Visual localization is the problem of estimating the position and orientation, \ie, the camera pose, from which the image was taken.   
Visual localization is a core component of intelligent systems such as self-driving cars~\cite{Heng2019ICRA} and other autonomous robots~\cite{Lim12CVPR}, augmented and virtual reality systems~\cite{Middelberg2014ECCV,Lynen2015RSS}, as well as of applications such as human performance capture~\cite{guzov2021human}.

In terms of pose accuracy, most of the current state-of-the-art in visual localization is structure-based~\cite{Sattler2017PAMI,Svarm2017PAMI,brachmann2020ARXIV,Brachmann2019ICCVa,Shotton2013CVPR,Schoenberger2018CVPR,Sarlin2019CVPR,Sarlin2020CVPR,HumenbergerX20Kapture,brejcha2020landscapear,Zeisl2015ICCV,Cavallari2019TPAMI,Cavallari20193DV}. 
These approaches establish 2D-3D correspondences between pixels in a query image and 3D points in the scene. 
The resulting 2D-3D matches can in turn be used to estimate the camera pose, \eg, by applying a minimal solver for the absolute pose problem~\cite{Fischler81CACM, Persson2018ECCV} inside a modern RANSAC implementation~\cite{Fischler81CACM,Lebeda2012BMVC,Chum09CVPR,barath2018graph,barath2019progressive,barath2020magsac++}.
The scene is either explicitly represented via a 3D model~\cite{Sattler2017PAMI,Sattler2015ICCV,Svarm2017PAMI,Zeisl2015ICCV,Li2010ECCV,Li2012ECCV,Irschara09CVPR,Sarlin2019CVPR,Sarlin2021CVPR,HumenbergerX20Kapture} or implicitly via the weights of a machine learning model~\cite{Brachmann2017CVPR,Brachmann2018CVPR,brachmann2020ARXIV,Brachmann2019ICCVa,Shotton2013CVPR,Valentin20163DV,Massiceti2017ICRA,Cavallari20193DV,Cavallari2017CVPR,Cavallari2019TPAMI}.

Methods that explicitly represent the scene via a 3D model have been shown to scale to city-scale~\cite{Svarm2017PAMI,Sattler2015ICCV,Zeisl2015ICCV} and beyond~\cite{Li2012ECCV}, while being robust to illumination, weather, and seasonal changes~\cite{Toft2020TPAMI,Sarlin2020CVPR,Sarlin2021CVPR,HumenbergerX20Kapture}.
These approaches typically use local features to establish the 2D-3D matches. 
The dominant 3D scene representation is a Structure-from-Motion (SfM)~\cite{Schoenberger2016CVPR,Agarwal2009ICCV,Snavely08IJCV} model. 
Each 3D point in these sparse point clouds was triangulated from local features found in two or more database images. 
To enable 2D-3D matching between the query image and the 3D model, each 3D point is associated with its corresponding local features. 
While such approaches achieve state-of-the-art results, they are rather \emph{inflexible}. 
Whenever a better type of local features becomes available, it is necessary to recompute the point cloud. 
Since the intrinsic calibrations and camera poses of the database images are available, it is sufficient to re-triangulate the scene rather than running SfM from scratch.
Still, computing the necessary feature matches between database images can be highly time-consuming. 

\begin{figure}[t]
    \centering
    \includegraphics[width=0.95\textwidth]{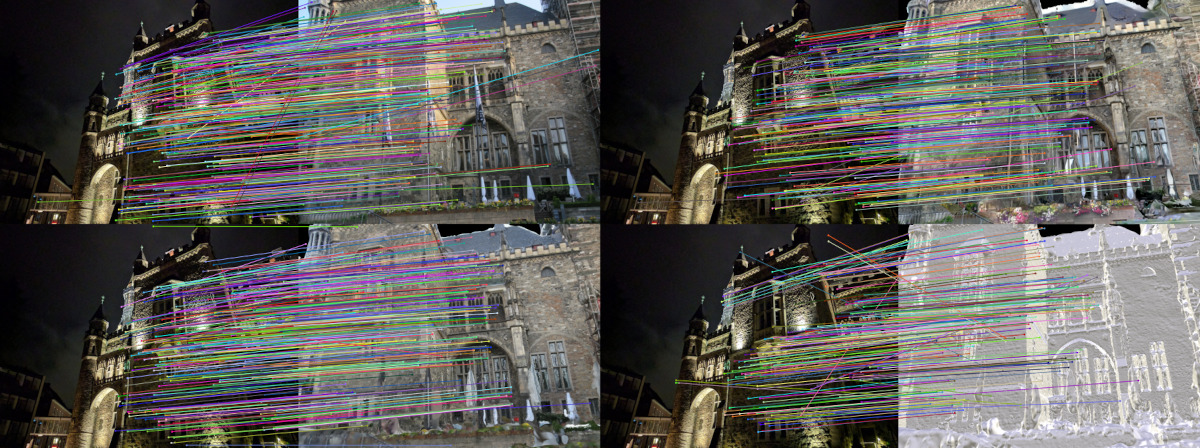}
    \caption{Modern learned features such as Patch2Pix~\cite{Zhou2021CVPR} are not only able to establish correspondences between real images (top-left), but are surprisingly good at matching between a real images and non-photo-realistic synthetic views (top-right: textured mesh, bottom-left: colored mesh, bottom-right: raw surface rendering). This observation motivates our investigation into using dense 3D meshes, rather than the Structure-from-Motion point clouds predominantly used in the literature}
    \label{fig:teaser}
\end{figure}

Often, it is possible to obtain a dense 3D model of the scene, \eg, in the form of a mesh obtained via multi-view stereo~\cite{Schoenberger2016ECCV,Kazhdan2013TOG}, from depth data, from LiDAR, or from other sources such as digital elevation models~\cite{brejcha2020landscapear,Tomesek2022CrossLocateCL}. 
Using a dense model instead of a sparse SfM point cloud offers more flexibility: 
rather than having to match features between database images to triangulate 3D scene points, one can simply obtain the corresponding 3D point from depth maps rendered from the model. 
Due to decades of progress in computer graphics research and development, even large 3D models can be rendered in less than a millisecond. 
Thus, feature matching and depth map rendering can both be done online without the need to pre-extract and store local features. 
This leads to the question whether one needs to store images at all or could render views of the model on demand. 
This in turn leads to the question how realistic these renderings need to be and thus which level of detail is required from the 3D models. %

This paper investigates using dense 3D models instead of sparse SfM point clouds for feature-based visual localization. 
Concretely, the paper makes the following contributions: 
(1) we discuss how to design a dense 3D model-based localization pipeline and contrast this system to standard hierarchical localization systems. 
(2) we show that a very simple version of the pipeline can already achieve state-of-the-art results when using the original images and a 3D model that accurately aligns with these images. 
Our mesh-based framework reduces overhead in testing local features and feature matchers for visual localization tasks compared to SfM point cloud-based methods. 
(3) we show interesting and promising results when using non-photo-realistic renderings of the meshes instead of real images in our pipeline. 
In particular, we show that existing features, applied out-of-the-box without fine-tuning or re-training, perform surprisingly well when applied on renderings of the raw 3D scene geometry without any colors or textures (\cf Fig.~\ref{fig:teaser}). 
We believe that this result is interesting as it shows 
that standard local features can be used to match images and purely geometric 3D models, \eg, laser or LiDAR scans. 
(4) our code and data are publicly available at \url{https://github.com/tsattler/meshloc\_release}.

\PAR{Related work.} 
One main family of state-of-the-art visual localization algorithms is based on local features~\cite{Sattler2017PAMI,Svarm2017PAMI,Li2012ECCV,Schoenberger2018CVPR,Sarlin2019CVPR,Sarlin2020CVPR,HumenbergerX20Kapture,brejcha2020landscapear,Zeisl2015ICCV,Sattler2015ICCV,Taira2018CVPR,Taira2019ICCV}. 
These approaches commonly represent the scene as a sparse SfM point cloud, where each 3D point was triangulated from features extracted from the database images. 
At test time, they establish 2D-3D matches between pixels in a query image and 3D points in the scene model using descriptor matching.  
In order to scale to large scenes and handle complex illumination and seasonal changes, %
a hybrid approach is often used~\cite{Irschara09CVPR,Sattler2012BMVC,Sarlin2018CORL,Taira2018CVPR,Taira2019ICCV,HumenbergerX20Kapture}: 
an image retrieval stage~\cite{Torii-CVPR15,Arandjelovic2016CVPR} is used to identify a small set of potentially relevant database images. 
Descriptor matching is then restricted to the 3D points visible in these images. 
We show that it is possible to achieve similar results using a mesh-based scene representation that allows researchers to more easily experiment with new types of features. 

An alternative to explicitly representing the 3D scene geometry via a 3D model is to implicitly store information about the scene in the weights of a machine learning model.
Examples include scene coordinate regression techniques~\cite{Shotton2013CVPR,Brachmann2017CVPR,Brachmann2018CVPR,brachmann2020ARXIV,Valentin20163DV,Cavallari20193DV,Cavallari2017CVPR,Cavallari2019TPAMI}, which regress 2D-3D matches rather than computing them via explicit descriptor matching, and absolute~\cite{Kendall2015ICCV,Kendall2017CVPR,Shavit2021ICCV,Walch2017ICCV,Moreau2021CORL} and relative pose~\cite{Ding2019ICCV,Balntas2018ECCV,Laskar2017ICCVW} regressors. 
Scene coordinate regressors achieve state-of-the-art results for small scenes~\cite{Brachmann2021ICCV}, but have not yet shown strong performance in more challenging scenes. 
In contrast, absolute and relative pose regressors are currently not (yet) competitive to feature-based methods~\cite{Sattler2019CVPR,Zhou2020ICRA}, even when using additional training images obtained via view synthesis~\cite{Moreau2021CORL,Ng2021ARXIV,Naseer2017IROS}. 

Ours is not the first work to use a dense scene representation. 
Prior work has used dense Multi-View Stereo~\cite{Shan-3DV14} and laser~\cite{Sibbing133DV,Taira2018CVPR,Taira2019ICCV,Naseer2017IROS,Ng2021ARXIV} point clouds as well as textured or colored meshes~\cite{brejcha2020landscapear,Mueller2019PIA,Zhang2020IJCV}. 
\cite{Shan-3DV14,Sibbing133DV,Naseer2017IROS,Ng2021ARXIV,Mueller2019PIA} render novel views of the scene to enable localization of images taken from viewpoints that differ strongly from the database images. 
Synthetic views of a scene, rendered from an estimated pose, can also be used for pose verification~\cite{Taira2018CVPR,Taira2019ICCV}. 
\cite{Mueller2019PIA,Moreau2021CORL,Zhang2020IJCV} rely on neural rendering techniques such as Neural Radiance Fields (NeRFs)~\cite{mildenhall2020nerf,mueller2022instant} or image-to-image translation~\cite{CycleGAN2017} while \cite{Shan-3DV14,Sibbing133DV,Naseer2017IROS,Zhang2020IJCV} rely on classical rendering techniques. 
Most related to our work are~\cite{brejcha2020landscapear,Zhang2020IJCV} as both use meshes for localization: 
given a rather accurate prior pose, provided manually, \cite{Zhang2020IJCV} render the scene from the estimated pose and match features between the real image and the rendering. 
This results in a set of 2D-3D matches used to refine the pose. 
While \cite{Zhang2020IJCV} start with poses close to the ground truth, we show that meshes can be used to localize images from scratch and describe a full pipeline for this task.  
While the city scene considered in~\cite{Zhang2020IJCV} was captured by images, \cite{brejcha2020landscapear} consider localization in mountainous terrain, where only few database images are available. 
As it is impossible to compute an SfM point cloud from the sparsely distributed database images, they instead use a textured digital elevation model as their scene representation. 
They train local features to match images and this coarsely textured mesh, whereas we use learned features without re-training or fine-tuning. 
While \cite{brejcha2020landscapear} focus on coarse localization (on the level of hundreds of meters or even kilometers), we show that meshes can be used for centimeter-accurate localization. 
Compared to these prior works, %
we provide a detailed ablation study investigating how model and rendering quality impact the localization accuracy. %

\section{Feature-based Localization via SfM Models}
\label{sec:method_sfm}
This section first reviews the general outline of state-of-the-art hierarchical structure-based localization pipelines. 
Sec.~\ref{sec:method} then describes how such a pipeline can be adapted when using a dense instead of a sparse scene representation.

\noindent \textbf{Stage 1: Image Retrieval.} 
Given a set of database images, this stage identifies a few relevant reference views for a given query. 
This is commonly done via nearest neighbor search with image-level descriptors~\cite{Torii2015CVPR,Arandjelovic2016CVPR,gordo2017end,revaud2019learning}.

\noindent \textbf{Stage 2: 2D-2D Feature Matching.} 
This stage establishes feature matches between the query image and the top-$k$ retrieved database images, which will be upgraded to 2D-3D correspondences in the next stage. 
It is common to use state-of-the-art learned local features~\cite{Sun2021CVPR,Zhou2021CVPR,DeTone2018CVPRWorkshops,Sarlin2020CVPR,revaud2019r2d2,Dusmanu2019CVPR}. 
Matches are established either by (exhaustive) feature matching, potentially followed by outlier filters such as Lowe's ratio test~\cite{Lowe04IJCV}, or using learned matching strategies~\cite{Sarlin2020CVPR,Rocco2018NeurIPS,Rocco2020ECCV,Zhou2021CVPR}. 

There are two representation choices for this stage: 
pre-compute the features for the database images or only store the photos and extract the features on-the-fly.
The latter requires less storage at the price of run time overhead. 
\Eg, storing SuperPoint~\cite{DeTone2018CVPRWorkshops} features for the Aachen Day-Night v1.1 dataset~\cite{Sattler2012BMVC,Sattler2018CVPR,Zhang2020IJCV} requires more than 25 GB while the images themselves take up only 7.5 GB (2.5 GB when reducing the image resolution to at most 800 pixels). 

\noindent \textbf{Stage 3: Lifting 2D-2D to 2D-3D Matches.} 
For the $i$-th 3D scene point $\mathbf{p}_i\in \mathbb{R}^3$, each SfM point cloud stores a set $\{(\mathcal{I}_{i_1}, \mathbf{f}_{i_1}), \cdots (\mathcal{I}_{i_n}, \mathbf{f}_{i_n})\}$ of (image, feature) pairs. 
Here, a pair $(\mathcal{I}_{i_j}, \mathbf{f}_{i_j})$ denotes that feature $\mathbf{f}_{i_j}$ in image $\mathcal{I}_{i_j}$ was used to triangulate the 3D point position $\mathbf{p}_i$. 
If a feature in the query image matches %
$\mathbf{f}_{i_j}$ in database image $\mathcal{I}_{i_j}$, it %
thus also matches %
$\mathbf{p}_i$. 
Thus, 2D-3D matches are obtained %
by looking up 3D points corresponding to matching database features.

\noindent \textbf{Stage 4: Pose Estimation.}
The last stage uses %
the resulting 2D-3D matches 
for camera pose estimation. 
It is common practice to use LO-RANSAC~\cite{Lebeda2012BMVC,Chum2002BMVC,Fischler81CACM} for robust pose estimation. 
In each iteration, a P3P solver~\cite{Persson2018ECCV} generates pose hypotheses from a minimal set of three 2D-3D matches.
Non-linear refinement over all inliers is used to optimize the pose, both inside and after LO-RANSAC.

\noindent \textbf{Covisibility filtering.} 
Not all matching 3D points might be visible together. 
It is thus common to use a covisibility filter~\cite{Sattler2012ECCV,Li2010ECCV,Li2012ECCV,Sarlin2018CORL}:  
a SfM reconstruction defines the so-called visibility graph $\mathcal{G}=((I,P),E)$~\cite{Li2010ECCV}, a bipartite graph where one set of nodes $I$ corresponds to the database images and the other set $P$ to the 3D points. 
$\mathcal{G}$ contains an edge between an image node and a point node if the 3D point has a corresponding feature in the image. 
A set $M = \{(\mathbf{f}_i, \mathbf{p}_i)\}$ of 2D-3D matches defines a subgraph $\mathcal{G}(M)$ of $\mathcal{G}$.
Each connected component of $\mathcal{G}(M)$ contains 3D points that are potentially visible together. 
Thus, pose estimation is done per connected component rather than over all matches~\cite{Sarlin2018CORL,Sattler2015ICCV}.

\section{Feature-based Localization without SfM Models}
\label{sec:method}
This paper aims to explore dense 3D scene models as an alternative to the sparse Structure-from-Motion (SfM) point clouds typically used in state-of-the-art feature-based visual localization approaches. 
Our motivation %
is three-fold: 

(1) dense scene models are more flexible than SfM-based representations: 
SfM point clouds are specifically build for a given type of feature. 
If we want to use another type, \eg, when evaluating %
the latest local feature from the literature, a new SfM point cloud needs to be build.  
Feature matches between the database images are required to triangulate SfM points. %
For medium-sized scenes, this matching process can take hours, for large scenes days or weeks. 
In contrast, once a dense 3D scene model is build, it can be used to directly provide the corresponding 3D point for (most of) the pixels in a database image by simply rendering a depth map. 
In turn, depth maps can be rendered highly efficiently when using 3D meshes, \ie, in a millisecond or less. 
Thus, there is only very little overhead when evaluating a new type of local features.

(2) dense scene models can be rather compact: 
at first glance, it seems that storing a dense model will be much less memory efficient than storing a sparse point cloud. However, our experiments show that we can achieve state-of-the-art results on the Aachen v1.1 dataset~\cite{Sattler2012BMVC,Sattler2018CVPR,Zhang2020IJCV} using depth maps generated by a model that requires only 47 MB. This compares favorably to the 87 MB required to store the 2.3M 3D points and 15.9M corresponding database indices (for co-visibility filtering) for the SIFT-based SfM model provided by the dataset.

(3) as mentioned in Sec.~\ref{sec:method_sfm}, storing the original images and extracting features on demand
requires less memory
compared to directly storing the features. 
One intriguing possibility of dense scene representations is thus to not store images at all but to use rendered views for feature matching. 
Since dense representations such as meshes can be rendered 
in a millisecond or less, this rendering step introduces little run time overhead. 
It can also help to further reduce memory requirements: 
\Eg, a textured model of the Aachen v1.1~\cite{Sattler2012BMVC,Sattler2018CVPR,Zhang2020IJCV} dataset requires around 837 MB compared to the more than 7 GB needed for storing the original database images (2.5 GB at reduced resolution). 
While synthetic images can also be rendered from sparse SfM point clouds~\cite{Song2020ECCV,Pittaluga2019CVPR}, these approaches are in our experience orders of magnitude slower than rendering a 3D mesh. 

The following describes the design choices one has when adapting the hierarchical localization pipeline from Sec.~\ref{sec:method_sfm} to using dense scene representations.

\noindent \textbf{Stage 1: Image Retrieval.} 
We focus on exploring using dense representations for obtaining 2D-3D matches and %
do not make any changes to the retrieval stage. %
Naturally, use additional rendered views can be used to improve the retrieval performance~\cite{Irschara09CVPR,Sibbing133DV,Mueller2019PIA}. 
As we are interested in comparing classical SfM-based and dense representations, we do not investigate this direction of research though.

\noindent \textbf{Stage 2: 2D-2D Feature Matching.} 
Algorithmically, there is no difference between matching features between real images and a real query image and a rendered view.
Both cases result in a set of 2D-2D matches that can be upgraded to 2D-3D matches in the next stage. 
As such, we do not modify this stage. 
We employ state-of-the-art learned local features~\cite{Sun2021CVPR,Zhou2021CVPR,DeTone2018CVPRWorkshops,Sarlin2020CVPR,revaud2019r2d2,Dusmanu2019CVPR} and matching strategies~\cite{Sarlin2020CVPR}. %
We do not re-train any of the local features. 
Rather, we are interested in determining how well these features work out-of-the-box for non-photo-realistic images for different degrees of non-photo-realism, \ie, textured 3D meshes, colored meshes where each vertex has a corresponding RGB color, and raw geometry without any color.

\noindent \textbf{Stage 3: Lifting 2D-2D to 2D-3D Matches.}
In an SfM point cloud, each 3D point $\mathbf{p}_i$ has multiple corresponding features $\mathbf{f}_{i_1}, \cdots ,\mathbf{f}_{i_n}$ from database images $\mathcal{I}_{i_1}, \cdots ,\mathcal{I}_{i_n}$. 
Since the 2D feature positions are subject to noise, $\mathbf{p}_i$ will not precisely project to any of its corresponding features. 
$\mathbf{p}_i$ is computed such that it minimizes the sum of squared reprojection errors to these features, thus averaging out the noise in the 2D feature 
positions.
If a query feature $\mathbf{q}$ matches to features $\mathbf{f}_{i_j}$ and $\mathbf{f}_{i_k}$ belonging to $\mathbf{p}_i$, we obtain a single 2D-3D match $(\mathbf{q}, \mathbf{p}_i)$.

When using a depth map obtained by rendering a dense model, each database feature $\mathbf{f}_{i_j}$ with a valid depth will have a corresponding 3D point $\mathbf{p}_{i_j}$. 
Each $\mathbf{p}_{i_j}$ will project precisely onto its corresponding feature, \ie, the noise in the database feature positions is directly propagated to the 3D points. 
This implies that even though $\mathbf{f}_{i_1}, \cdots ,\mathbf{f}_{i_n}$ are all noisy measurements of the same physical 3D point, the corresponding model points $\mathbf{p}_{i_1}, \cdots ,\mathbf{p}_{i_n}$ will all be (slightly) different. 
If a query feature $\mathbf{q}$ matches to features $\mathbf{f}_{i_j}$ and $\mathbf{f}_{i_k}$, we thus obtain multiple (slightly) different 2D-3D matches $(\mathbf{q}, \mathbf{p}_{i_j})$ and $(\mathbf{q}, \mathbf{p}_{i_k})$. 

There are two options to handle the resulting multi-matches: 
(1) we \textbf{simply use all individual matches}. 
This strategy is extremely simple to implement, but can also produce a large number of matches. 
For example, when using the top-50 retrieved images, each query feature $\mathbf{q}$ can produce up to 50 2D-3D correspondences. 
This in turn slows down RANSAC-based pose estimation. 
In addition, it can bias the pose estimation process towards finding poses that are consistent with features that produce more matches. 

(2) we \textbf{merge multiple 2D-3D matches into a single 2D-3D match}: 
given a set $\mathcal{M}(\mathbf{q}) = \{(\mathbf{q}, \mathbf{p}_{i})\}$ of 2D-3D matches obtained for a query feature $\mathbf{q}$, we estimate a single 3D point $\mathbf{p}$, resulting in a single 2D-3D correspondence $(\mathbf{q}, \mathbf{p})$.   
Since the set $\mathcal{M}(\mathbf{q})$ can contain wrong matches, we first try to find a consensus set using the database features $\{\mathbf{f}_i\}$ corresponding to the matching points. 
For each matching 3D point $\mathbf{p}_{i}$, we measure the reprojection error \wrt to the database features and count the number of features for which the error is within a given threshold. 
The point with the largest number of such inliers
\footnote{We actually optimize a robust MSAC-like cost function~\cite{Lebeda2012BMVC} not the number of inliers.}
is then refined by optimizing its sum of squared reprojection errors \wrt the inliers. 
If there is no point $\mathbf{p}_i$ with at least two inliers, we keep all matches from $\mathcal{M}(\mathbf{q})$. 
This approach thus aims at averaging out the noise in the database feature detections to obtain more precise 3D point locations.

\noindent \textbf{Stage 4: Pose Estimation.} 
Given a set of 2D-3D matches, we follow the same approach as in Sec.~\ref{sec:method_sfm} for camera pose estimation. 
However, we need to adapt covisibility filtering and introduce a simple position averaging approach as a post-processing step after RANSAC-based pose estimation.

\noindent \textbf{Covisibility filtering.}
Dense scene representations do not directly provide the co-visibility relations encoded in the visibility graph $\mathcal{G}$ and we want to avoid computing matches between database images.
Naturally, one could compute visibility relations between views using their depth maps.
However, this approach is computationally expensive.
A more efficient alternative is to define the visibility graph on-the-fly via shared matches with query features:
the 3D points visible in views $\mathcal{I}_i$ and $\mathcal{I}_j$ are deemed co-visible if there exists at least one pair of matches $(\mathbf{q},\mathbf{f}_i)$, $(\mathbf{q},\mathbf{f}_j)$ between a query feature $\mathbf{q}$ and features $\mathbf{f}_i \in \mathcal{I}_i$ and $\mathbf{f}_j \in \mathcal{I}_j$.
In other words, the 3D points from two images are considered co-visible if at least one feature in the query image matches to a 3D point from each image.

Naturally, the 2D-2D matches (and the corresponding 2D-3D matches) define a  set of connected components and we can perform pose estimation per component.
However, the visibility relations computed on the fly are an approximation to the visibility relations encoded in $\mathcal{G}$:
images $\mathcal{I}_i$ and $\mathcal{I}_j$ might not share 3D points, but can observe the same 3D points as image $\mathcal{I}_k$.
In $\mathcal{G}(M)$, the 2D-3D matches found for images $\mathcal{I}_i$ and $\mathcal{I}_j$ thus belong to a single connected component.
In the on-the-fly approximation, this connection might be missed, \eg, if image $\mathcal{I}_k$ is not among the top-retrieved images.
Covisibility filtering using the on-the-fly approximation might thus be too aggressive, resulting in an over-segmentation of the set of matches and a drop in localization performance. 

\noindent \textbf{Position averaging.} 
The output of pose estimation approach is a camera pose $\mathtt{R}$, $\mathbf{c}$ and the 2D-3D matches that are inliers to that pose. 
Here, $\mathtt{R} \in \mathbb{R}^{3\times 3}$ is the rotation from global model coordinates to camera coordinates while $\mathbf{c}\in \mathbb{R}^3$ is the position of the camera in global coordinates.  
In our experience, the estimated rotation is often more accurate than the estimated position. 
We thus use a simple scheme to refine the position $\mathbf{c}$: 
we center a volume of side length $2 \cdot d_\text{vol}$ around the position $\mathbf{c}$. 
Inside the volume, we regularly sample new positions with a step size $d_\text{step}$ in each direction. 
For each such position $\mathbf{c}_i$, we count the number $I_i$ of inliers to the pose $\mathtt{R}$, $\mathbf{c}_i$ and obtain a new position estimate $\mathbf{c}'$ as the weighted average $\mathbf{c}' = \frac{1}{\sum_i I_i} \sum_i I_i \cdot \mathbf{c}_i$. 
Intuitively, this approach is a simple but efficient way to handle poses with larger position uncertainty: 
for these poses, there will be multiple positions with a similar number of inliers and the resulting position $\mathbf{c}'$ will be closer to their average rather than the position with the largest number of inliers (which might be affected by noise in the features and 3D points). 
Note that this averaging strategy is not tied to using a dense scene representations.

\section{Experimental Evaluation}
\label{sec:experiments}
We evaluate the localization pipeline described in Sec.~\ref{sec:method} on two publicly available datasets commonly used to evaluate visual localization algorithms, Aachen Day-Night v1.1~\cite{Sattler2012BMVC,Sattler2018CVPR,Zhang2020IJCV} and 12 Scenes~\cite{Valentin20163DV}. 
We use the Aachen Day-Night dataset to study the importance (or lack thereof) of the different components in the pipeline described in Sec.~\ref{sec:method}. 
Using the original database images, we evaluate the approach using multiple learned local features~\cite{Sun2021CVPR, Sarlin2020CVPR, Zhou2021CVPR, Wang2020ARXIV, revaud2019r2d2} and 3D models of different levels of detail. 
We show that the proposed approach can reach state-of-the-art performance compared to the commonly used SfM-based scene representations. 
We further study using renderings instead of real images to obtain the 2D-2D matches in Stage 2 of the pipeline, using 3D meshes of different levels of quality and renderings of different levels of detail. 
A main result is that modern features are robust enough to match real photos against non-photo-realistic renderings of raw scene geometry, even though they were never trained for such a scenario, resulting in surprisingly accurate pose estimates.

\begin{figure*}[t!]%
    \centering
    \includegraphics[width=0.95\columnwidth]{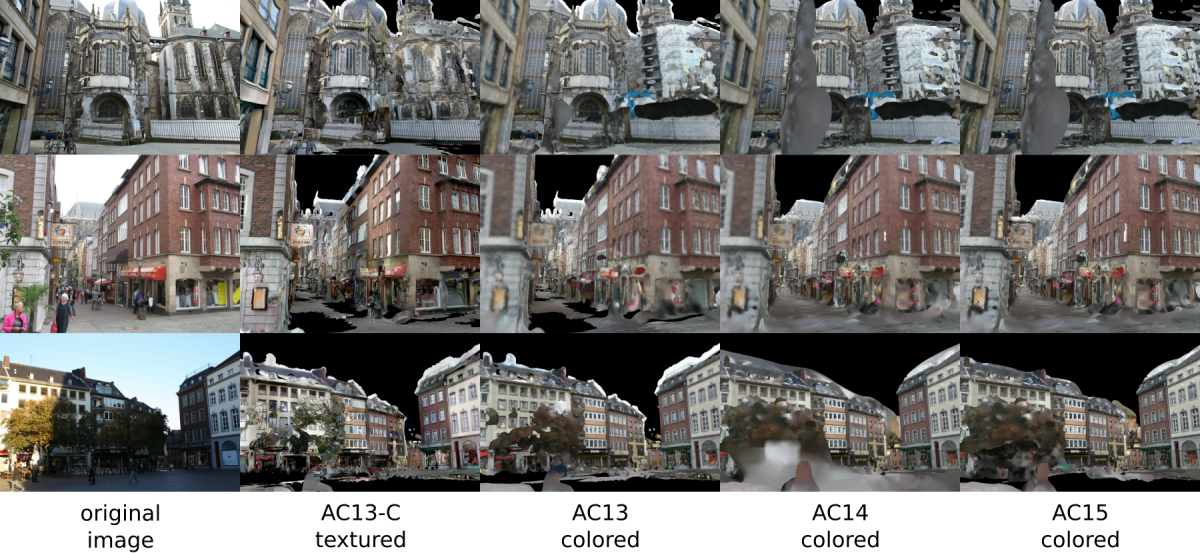} %
    \caption{\textbf{Examples of colored/textured renderings  from the Aachen Day-Night v1.1 dataset}~\cite{Sattler2012BMVC,Sattler2018CVPR,Zhang2020IJCV}. We use meshes of different levels of detail (from coarsest to finest: \emph{AC13-C}, \emph{AC13}, \emph{AC14}, and \emph{AC15}) and different rendering styles: a textured 3D model (only for \emph{AC13-C}) and meshes with per-vertex colors (\emph{colored}). For reference, the leftmost column shows the corresponding original database image.}%
    \label{fig:renderings_color_ac}
\end{figure*}
    
\begin{figure*}[t!]%
    \centering
    \includegraphics[width=1.0\textwidth]{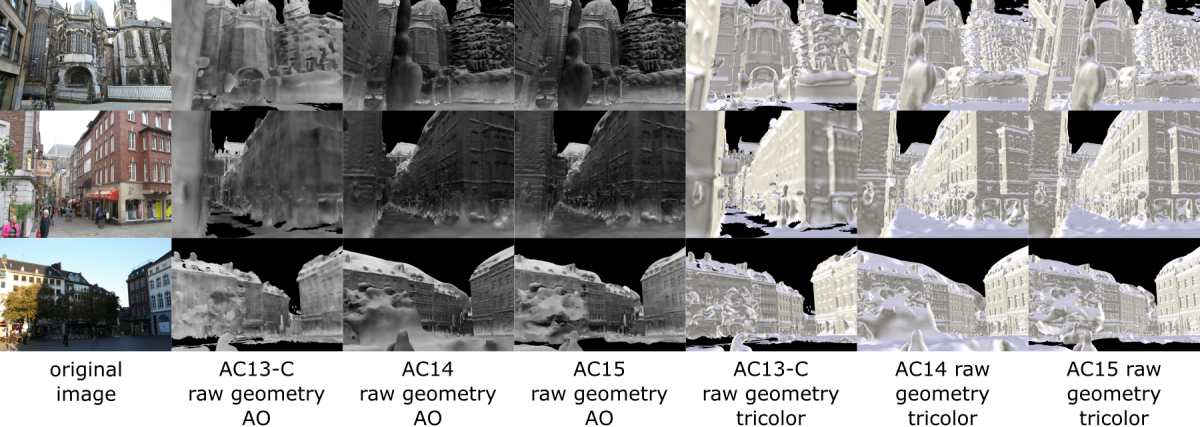} %
    \caption{\textbf{Example of raw geometry renderings for the Aachen Day-Night v1.1 dataset}~\cite{Sattler2012BMVC,Sattler2018CVPR,Zhang2020IJCV}. We use different rendering styles to generate synthetic views of the \emph{raw scene geometry}: \emph{ambient occlusion}~\cite{Zhukov1998AnAL} (AO) and illumination from three colored lights (\emph{tricolor}). The leftmost column shows the corresponding original database image.}%
    \label{fig:renderings_raw_ac}
\end{figure*}

\begin{figure*}[t!]%
    \centering
    \includegraphics[width=1.0\textwidth]{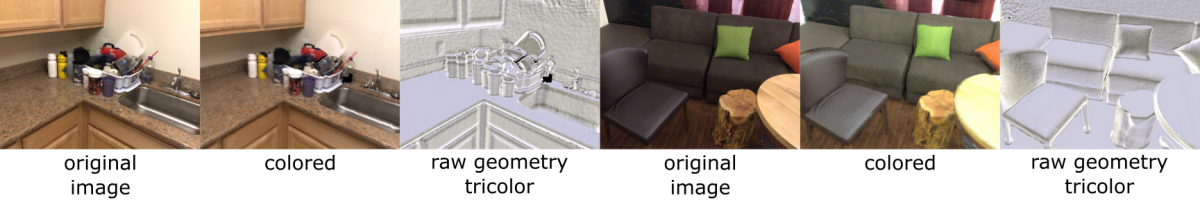} %
    \caption{\textbf{Example renderings for the 12 scenes dataset}~\cite{Valentin20163DV}.}
    \label{fig:renderings_12_scenes}
\end{figure*}

\noindent \textbf{Datasets.} %
The Aachen Day-Night v1.1 dataset~\cite{Sattler2012BMVC,Sattler2018CVPR,Zhang2020IJCV} contains 6,697 database images captured in the inner city of Aachen, Germany. 
All database images were taken under daytime conditions over %
multiple months. 
The dataset also containts 824 daytime and 191 nighttime query images captured with multiple smartphones.
We use only the more challenging night subset for evaluation.

To create dense 3D models for the Aachen Day-Night dataset, we %
use Screened Poisson Surface Reconstruction (SPSR)~\cite{Kazhdan2013TOG} to create 3D meshes from Multi-View Stereo%
~\cite{Schoenberger2016ECCV} point clouds. %
We generate meshes of different levels of quality by varying the depth parameter of SPSR, controlling the maximum resolution of the Octree that is used to generate the final mesh. 
Each of the resulting meshes, AC13, AC14, and AC15 
(corresponding to depths 13, 14, and 15, with larger depth values corresponding to more detailed models), has an RGB color associated to each of its vertices. 
We further generate a compressed version of AC13, denoted as AC13-C, using \cite{jakob2015instant} and texture it using \cite{waechter2014let}. 
Fig.~\ref{fig:renderings_color_ac} shows examples.

The 12 Scenes dataset~\cite{Valentin20163DV} consists of 12 room-scale indoor scenes captured using RGB-D cameras, with ground truth poses created using RGB-D SLAM~\cite{Dai2017TOG}.  
Each scene provides RGB-D query images, but we only use the RGB part for evaluation. 
The dataset further provides the colored meshes reconstructed using \cite{Dai2017TOG}, where each vertex is associated with an RGB color, which we use for our experiments. 
Compared to the Aachen Day-Night dataset, the 12 Scenes dataset is ``easier" in the sense that it only contains images taken by a single camera that is not too far away from the scene and under constant illumination conditions. 
Fig.~\ref{fig:renderings_12_scenes} shows example renderings.

For both datasets, we render depth maps and images from the meshes using an OpenGL-based rendering pipeline~\cite{Waechter2017TOG}. 
Besides rendering colored and textured meshes, we also experiment with raw geometry rendering. 
In the latter case, no colors or textures are stored, which reduces memory requirements. 
In order to be able to extract and match features, we rely on shading cues. 
We evaluate two shading strategies for the raw mesh geometry rendering: 
the first uses ambient occlusion~\cite{Zhukov1998AnAL} (AO) pre-computed in MeshLab~\cite{Cignoni2008MeshLabAO}. 
The second one uses three colored light sources (tricolor)
(Sec.~\ref{sec:implementation}). %
Figs.~\ref{fig:renderings_raw_ac} and~\ref{fig:renderings_12_scenes} show example renderings. 
Statistics about the meshes and rendering times can be found in Tab.~\ref{tab:mesh_list} for Aachen and in 
Tab.~\ref{tab:mesh_list_12_scenes} for 12 Scenes. %

This paper focuses on dense scene representations based on meshes. 
Hence, we refer to the pipeline from Sec.~\ref{sec:method} as MeshLoc. 
A more modern dense scene representations are %
NeRFs~\cite{mildenhall2020nerf,Barron2021MipNeRFAM, MartinBrualla2021NeRFIT, Tancik2022BlockNeRFSL, Wang2021IBRNetLM}. %
Preliminary experiments with a recent NeRF implementation~\cite{mueller2022instant} resulted in realistic renderings for the 12 Scenes dataset~\cite{Valentin20163DV}. 
Yet, we were not able to obtain useful depth maps. 
We attribute this to the fact that the NeRF representation can compensate for noisy occupancy estimates via the predicted color~\cite{Oechsle2021ICCV}.
We thus leave a more detailed exploration of neural rendering strategies for future work.
At the moment we use well-matured OpenGL-based rendering on standard 3D meshes, which is optimized for GPUs and achieves very fast rendering times (see Tab.~\ref{tab:mesh_list}). %
See Sec.~\ref{sec:nerf} for further discussion on use of NeRFs. %

\begin{table*}[t!]
    \begin{center}
    \begin{minipage}{\textwidth}
    \caption{Statistics for the 3D meshes used for experimental evaluation as well as rendering times for different rendering styles and resolutions}
    \label{tab:mesh_list}
    \scriptsize{
    \renewcommand{\arraystretch}{0.8}
    \begin{tabular*}{\textwidth}{@{\extracolsep{\fill}}cllrrrrr@{\extracolsep{\fill}}}
        \hline\noalign{\smallskip}
        
        & & & & & & \multicolumn{2}{c}{Render time $[\mu s]$} \\ 
        \cline{7-8}\noalign{\smallskip}
        \multicolumn{2}{l}{Model} & Style & Size [MB] & Vertices & Triangles & 800 px & full res. \\
        \noalign{\smallskip}
        \hline
        \noalign{\smallskip}
        \parbox[t]{1mm}{\multirow{9}{*}{\rotatebox[origin=c]{-90}{Aachen v1.1}}} & AC13-C & textured & 645 & $1.4 \cdot 10^{6}$ & $2.4 \cdot 10^{6}$ & 1143 & 1187 \\
        & AC13-C & tricolor & 47 & $1.4 \cdot 10^{6}$ & $2.4 \cdot 10^{6}$ & 115 & 219 \\
        & AC13 & colored & 617 & $14.8 \cdot 10^{6}$ & $29.3 \cdot 10^{6}$ & 92 & 140 \\
        & AC13 & tricolor & 558 & $14.8 \cdot 10^{6}$ & $29.3 \cdot 10^{6}$ & 97 & 152 \\
        & AC14 & colored & 1234 & $29.4 \cdot 10^{6}$ & $58.7 \cdot 10^{6}$ & 100 & 139 \\
        & AC14 & tricolor & 1116 & $29.4 \cdot 10^{6}$ & $58.7 \cdot 10^{6}$ & 93 & 205 \\
        & AC15 & colored & 2805 & $66.8 \cdot 10^{6}$ & $133.5 \cdot 10^{6}$ & 98 & 137 \\
        & AC15 & tricolor & 2538 & $66.8 \cdot 10^{6}$ & $133.5 \cdot 10^{6}$ & 97 & 160 \\
        \hline
    \end{tabular*}
    }
    \end{minipage}
    \end{center}
\end{table*}

\noindent \textbf{Experimental setup.} 
We evaluate multiple learned local features and matching strategies: %
SuperGlue~\cite{Sarlin2020CVPR} (SG) first extracts and matches SuperPoint~\cite{DeTone2018CVPRWorkshops} features before applying a learned matching strategy to filter outliers. 
While SG is based on explicitly detecting local features, LoFTR~\cite{Sun2021CVPR} and Patch2Pix~\cite{Zhou2021CVPR} (P2P) densely match descriptors between pairs of images and extract matches from the resulting correlation volumes. 
Patch2Pix+SuperGlue (P2P+SG) uses the match refinement scheme from \cite{Zhou2021CVPR} to refine the keypoint coordinates of SuperGlue matches. 
For merging 2D-3D matches, we follow \cite{Zhou2021CVPR} and cluster 2D match positions in the query image to handle the fact that P2P and P2P+SG do not yield repeatable keypoints.
Sec.~\ref{sec:aachen} %
provides additional results with R2D2~\cite{revaud2019r2d2} and CAPS~\cite{Wang2020ARXIV} descriptors.

Following \cite{Sattler2018CVPR,Toft2020TPAMI,Jafarzadeh2021ICCV,Shotton2013CVPR,Valentin20163DV,Brachmann2021ICCV}, we report the percentage of query images localized within $X$ meters and $Y$ degrees of their respective ground truth poses.

We use the LO-RANSAC~\cite{Chum2002BMVC,Lebeda2012BMVC} implementation from %
PoseLib~\cite{PoseLib} %
with a robust Cauchy loss for non-linear refinement 
(Sec.~\ref{sec:implementation}). %

\begin{table}[t!]
    \begin{center}
    \begin{minipage}{\textwidth}
    \caption{Ablation study on the Aachen Day-Night v1.1 dataset~\cite{Sattler2012BMVC,Sattler2018CVPR,Zhang2020IJCV} using real images at reduced (max. side length 800 px) and full resolution (res.), and depth maps rendered using the AC13 model. We evaluate different strategies for obtaining 2D-3D matches (using all individual matches (I), merging matches (M), or triangulation (T)), with and without covisibility filtering (C), and with and without position averaging (PA) for various local features. We report the percentage of nighttime query images localized within 0.25m and 2$^\circ$ / 0.5m and 5$^\circ$ / 5m and 10$^\circ$ of the ground truth pose. For reference, we also report the corresponding results (from visuallocaliztion.net) obtained using  SfM-based representations (last row). Best results per feature are marked in bold}
    \label{tab:ablation_study_real_images_800}
    \scriptsize{
    \setlength\tabcolsep{4pt}
    \renewcommand{\arraystretch}{0.8}
    \begin{tabular*}{\textwidth}{@{\extracolsep{\fill}}cccccccc@{\extracolsep{\fill}}}
        \hline\noalign{\smallskip}
        
         & 2D- & & & SuperGlue & LoFTR & Patch2Pix & Patch2Pix\\ 
        res. & \ 3D & C & PA &  (SG)~\cite{Sarlin2020CVPR} & \cite{Sun2021CVPR} &  \cite{Zhou2021CVPR} &  + SG~\cite{Zhou2021CVPR} \\
        
        \noalign{\smallskip}
        \hline
        \noalign{\smallskip}
        
        \multirow{8}{*}{800} & I & & &
        72.8/\textbf{93.2}/99.0 & 77.0/92.1/\textbf{99.5} & 70.7/89.0/95.3 & 72.3/91.6/100.0 \\
         & I & \checkmark & & 
        72.3/92.7/99.0 & 76.4/92.1/\textbf{99.5} & 72.3/\textbf{91.1}/97.4 & 73.3/91.1/99.5 \\
         & I & & \checkmark & 
        74.3/93.2/99.0 & \textbf{78.5}/\textbf{93.2}/\textbf{99.5} & {73.8}/89.5/95.3 & 73.8/92.1/99.5  \\
         & I & \checkmark & \checkmark & 
        73.3/92.1/99.0 & 77.5/92.7/\textbf{99.5} & 73.3/\textbf{91.1}/97.4 & 73.8/91.1/99.5  \\
         & M & & \checkmark & 
        75.4/92.7/99.5 & 77.0/92.7/\textbf{99.5} & 70.7/89.5/96.3 & 73.8/92.7/99.5  \\
         & M & \checkmark & \checkmark & 
        75.4/91.6/99.5 & 75.4/92.1/99.5 & 69.6/89.0/97.4 & 72.8/\textbf{93.2}/\textbf{100.0}  \\
         & T & & \checkmark & 
        72.3/90.1/97.9 & 73.3/90.6/98.4 & 63.9/83.8/94.8 & 70.7/90.6/97.4  \\
         & T & \checkmark & \checkmark & 
        71.7/89.5/97.9 & 73.8/90.6/98.4 & 62.8/82.2/94.2 & 72.3/90.6/97.9  \\
        
        \hline
        \noalign{\smallskip}
        
        full & I & & \checkmark & \textbf{77.0}/92.1/99.0 & & \textbf{74.3}/90.1/96.3 & {74.3}/{92.1}/{99.5} \\
        
        \hline
        \noalign{\smallskip}
        
        SfM &  &  & & \textbf{77.0}/90.6/\textbf{100.0} & \textbf{78.5}/90.6/99.0 & 72.3/88.5/\textbf{97.9} & \textbf{78.0}/90.6/99.0 \\
        
        \hline
    \end{tabular*}
    }
    \end{minipage}
    \end{center}
\end{table}

\PAR{Experiments on Aachen Day-Night.}
We first study the importance of the individual components of the MeshLoc pipeline. %
We evaluate the pipeline on real database images and on rendered views of different level of detail and quality. 
For the retrieval stage, we follow the literature~\cite{Sarlin2019CVPR,Sarlin2020CVPR,Sun2021CVPR,Zhou2021CVPR} and use the top-50 retrieved database images / renderings based on  NetVLAD~\cite{Arandjelovic2016CVPR} descriptors extracted from the real database and query images. 

\noindent \textbf{Studying the individual components of MeshLoc.} 
Tab.~\ref{tab:ablation_study_real_images_800} presents an ablation study for the individual components of the MeshLoc pipeline from Sec.~\ref{sec:method}. 
Namely, we evaluate combinations of using all available individual 2D-3D matches (I) or merging 2D-3D matches for each query features (M), using the approximate covisibility filter (C), and position averaging (PA). 
We also compare a baseline that triangulates 3D points from 2D-2D matches between the query image and multiple database images (T) rather than using depth maps. 

As can be seen from the results of using down-scaled images 
(with maximum side length of 800 px),
using 3D points obtained from the AC13 model depth maps typically leads to better results than triangulating 3D points. 
For triangulation, we only use database features that match to the query image. 
Compared to an SfM model, where features are matched between database images, this leads to fewer features that are used for triangulation per point and thus to less accurate points. 
Preliminary experiments confirmed that, as expected, the gap between using the 3D mesh and triangulation grows when retrieving fewer database images. 
Compared to SfM-based pipelines, which use covisibility filtering before RANSAC-based pose estimation, we observe that covisibility filtering typically decreases the pose accuracy of the MeshLoc pipeline due to its approximate nature. 
Again, preliminary results showed that the effect is more pronounced when using fewer retrieved database images (as the approximation becomes coarser). In contrast, position averaging (PA) typically gives a (slight) accuracy boost. 
We further observe that the simple baseline that uses all individual matches (I) often leads to similar or better results compared to merging 2D-3D matches (M). 
In the following, we thus focus on a simple version of MeshLoc, which uses individual matches (I) and PA, but not covisibility filtering. 

\noindent \textbf{Comparison with SfM-based representations.} Tab.~\ref{tab:ablation_study_real_images_800} also evaluates the simple variant of MeshLoc on full-resolution images and compares MeshLoc against the corresponding SfM-based results from \href{https://www.visuallocalization.net/}{visuallocalization.net}. 
Note that we did not evaluate LoFTR on the full-resolution images due to the memory constraints of our GPU (NVIDIA GeForce RTX 3060, 12 GB RAM).  
The simple MeshLoc variant performs similarly well or slightly better than its SfM-based counterparts, with the exception of the finest pose threshold (0.25m, 2$^\circ$) for Patch2Pix+SG. 
This is despite the fact that SfM-based pipelines are significantly more complex and use additional information (feature matches between database images) that are expensive to compute. 
Moreover, MeshLoc requires less memory at only a small run time overhead 
(see Sec.~\ref{sec:storage_and_time}). %
Given its simplicity and ease of use, we thus believe that MeshLoc will be of interest to the community as it allows researchers to more easily prototype new features.

\begin{table}[t!]
    \begin{center}
    \begin{minipage}{\textwidth}
    \caption{Ablation study on the Aachen Day-Night v1.1 dataset~\cite{Sattler2012BMVC,Sattler2018CVPR,Zhang2020IJCV} using real images at reduced resolution (max. 800 px) and full resolution with depth maps rendered from 3D meshes of different levels of detail (\cf Tab.~\ref{tab:mesh_list}). We use a simple MeshLoc variant that uses individual matches and position averaging, but no covis. filtering }
    \label{tab:ablation_study_real_images_800_meshes}
    \scriptsize{
    \setlength\tabcolsep{4pt}
    \renewcommand{\arraystretch}{0.8}
    \begin{tabular*}{\textwidth}{@{\extracolsep{\fill}}cccccc@{\extracolsep{\fill}}}
        \hline\noalign{\smallskip}
        
        Feature & res. & AC13-C & AC13 & AC14 & AC15\\ 
        
        \noalign{\smallskip}
        \hline
        \noalign{\smallskip}
        
        SuperGlue~\cite{Sarlin2020CVPR} & \multirow{4}{*}{800} & 74.3/92.7/99.5  & 
        74.3/93.2/99.0  & 71.7/91.6/99.0  & 72.8/92.7/99.5\\
        LoFTR~\cite{Sun2021CVPR} & & 77.5/92.7/99.5  & 
        78.5/93.2/99.5  & 76.4/92.1/99.5  & 78.0/92.7/99.5\\
        Patch2Pix~\cite{Zhou2021CVPR} & & 71.7/88.0/95.3  & 
        73.8/89.5/95.3  & 67.0/85.9/95.8  & 72.3/89.0/96.3\\
        Patch2Pix+SG~\cite{Zhou2021CVPR, Sarlin2020CVPR} & & 74.9/92.1/99.5  & 
        73.8/92.1/99.5  & 73.8/90.1/99.0  & 75.4/91.1/99.5 \\

        \noalign{\smallskip}
        \hline
        \noalign{\smallskip}
        
        SuperGlue~\cite{Sarlin2020CVPR} & \multirow{3}{*}{full} & 77.0/92.1/99.5  & 
        77.0/92.1/99.0  & 75.4/91.1/99.0  & 76.4/92.1/99.0\\
        Patch2Pix~\cite{Zhou2021CVPR} & & 74.3/90.1/96.9  & 
        74.3/90.1/96.3  & 71.2/86.9/95.3  & 72.3/88.0/96.9\\
        Patch2Pix+SG~\cite{Zhou2021CVPR, Sarlin2020CVPR} & & 73.3/92.1/99.5  & 
        74.3/92.1/99.5  & 73.3/91.1/99.5  & 74.3/92.7/99.5 \\
        
        \hline
    \end{tabular*}
    }
    \end{minipage}
    \end{center}
\end{table}

\noindent \textbf{Mesh level of detail.} Tab.~\ref{tab:ablation_study_real_images_800_meshes} shows results obtained when using 3D meshes of different levels of detail (\cf Tab.~\ref{tab:mesh_list}). 
The gap between using the compact AC13-C model (47 MB to store the raw geometry) and the larger AC13 model (558 MB for the raw geometry) is rather small. 
While AC14 and AC15 offer more detailed geometry, they also contain artefacts in the form of blobs of geometry 
(Sec.~\ref{sec:qualitative}). %

Note that we did not optimize these models (besides parameter adjustments) %
and leave experiments with more accurate 3D models for future work. 
Overall the level of detail does not seem to be critical for MeshLoc. 

\begin{table}[t!]
    \begin{center}
    \begin{minipage}{\textwidth}
    \caption{Ablation study on the Aachen Day-Night v1.1 dataset~\cite{Sattler2012BMVC,Sattler2018CVPR,Zhang2020IJCV} using images rendered at reduced resolution (max. 800 px) from 3D meshes of different levels of detail (\cf Tab.~\ref{tab:mesh_list}) and different rendering types (textured / colored, raw geometry with ambient occlusion (AO), raw geometry with tricolor shading (tricolor)). For reference, the rightmost column shows results obtained with real images on AC13. MeshLoc %
    uses individual matches and position averaging, but no covisibility filtering }
    \label{tab:ablation_study_rendered}
    \scriptsize{
    \setlength\tabcolsep{4pt}
    \renewcommand{\arraystretch}{0.8}
    \begin{tabular*}{\textwidth}{@{\extracolsep{\fill}}ccccc@{\extracolsep{\fill}}}
        \hline\noalign{\smallskip}
        
         AC13-C: & textured & AO & tricolor & real \\ \hline
        SuperGlue~\cite{Sarlin2020CVPR} & 72.3/91.1/99.0 & 0.5/3.1/24.6 & 7.3/23.0/53.9 & 74.3/92.7/99.5 \\
        Patch2Pix+SG~\cite{Zhou2021CVPR, Sarlin2020CVPR} & 70.7/90.6/99.5 & 1.0/4.2/27.7 & 9.4/25.1/57.6 & 74.9/92.1/99.5 \\

        \hline
        \noalign{\smallskip}
        
        AC13: & colored & AO & tricolor & real \\ \hline
        SuperGlue~\cite{Sarlin2020CVPR} & 68.1/90.1/97.4 & 6.3/19.9/45.5 & 22.0/50.8/74.3 & 74.3/93.2/99.0\\
        Patch2Pix+SG~\cite{Zhou2021CVPR, Sarlin2020CVPR} & 71.7/91.1/97.9 & 6.8/26.2/49.2 & 23.0/55.0/78.5 & 73.8/92.1/99.5 \\

        \hline
        \noalign{\smallskip}

        AC14: & colored & AO & tricolor & real \\ \hline
        SuperGlue~\cite{Sarlin2020CVPR} & 70.2/90.1/96.3 & 23.6/44.5/63.9 & 33.0/65.4/79.1 & 71.7/91.6/99.0 \\
        Patch2Pix+SG~\cite{Zhou2021CVPR, Sarlin2020CVPR} & 72.3/92.1/96.9 & 26.7/48.2/68.1 & 39.3/68.6/80.6 & 73.8/90.1/99.0\\

        \hline
        \noalign{\smallskip}
        
        AC15: & colored & AO & tricolor & real \\ \hline
        SuperGlue~\cite{Sarlin2020CVPR} & 75.4/89.5/98.4 & 24.1/47.1/63.4 & 37.2/60.7/77.5 & 72.8/92.7/99.5\\
        Patch2Pix+SG~\cite{Zhou2021CVPR, Sarlin2020CVPR} & 72.8/92.1/98.4 & 25.1/51.3/70.2 & 40.3/66.0/80.1 & 75.4/91.1/99.5\\
        \hline
    \end{tabular*}
    }
    \end{minipage}
    \end{center}
\end{table}

\noindent \textbf{Using rendered instead of real images.} 
Next, we evaluate the MeshLoc pipeline using synthetic images rendered from the poses of the database images instead of real images. 
Tab.~\ref{tab:ablation_study_rendered} shows results for 
various rendering settings, resulting in different levels of realism for the synthetic views. 
We focus on SuperGlue~\cite{Sarlin2020CVPR} and Patch2Pix + SuperGlue~\cite{Zhou2021CVPR,Sarlin2020CVPR}. 
LoFTR performed similarly well or %
better than both on textured and colored renderings, but %
worse when rendering raw geometry 
(Tab.~\ref{tab:ablation_study_rendered_supp1} and~\ref{tab:ablation_study_rendered_supp2}).%

As Tab.~\ref{tab:ablation_study_rendered} shows, the pose accuracy gap between using real images and textured / colored renderings is rather small. 
This shows that advanced neural rendering techniques, \eg, NeRFs~\cite{mildenhall2020nerf}, have only a limited potential to improve the results. 
Rendering raw geometry results in significantly reduced performance since neither SuperGlue nor Patch2Pix+SG were trained on this setting.
AO renderings lead to worse results compared to the tricolor scheme as the latter produces more sharp details (\cf Fig.~\ref{fig:renderings_raw_ac}). 
Patch2Pix+SuperGlue outperforms SuperGlue as it refines the keypoint detections used by SuperGlue on a per-match-basis~\cite{Zhou2021CVPR}, resulting in more accurate 2D positions and reducing the bias between positions in real and rendered images. 
Still, the results for the coarsest threshold (5m, 10$^\circ$) are surprisingly competitive. 
This indicates that there is quite some potential in matching real images against renderings of raw geometry, \eg, for using dense models obtained from non-image sources (laser, LiDAR, depth, \etc) for visual localization. 
Naturally, having more geometric detail leads to better results as it produces more fine-grained details in the renderings. 

\PAR{Experiments on 12 Scenes.}
The meshes provided by the 12 Scenes dataset~\cite{Valentin20163DV} come from RGB-D SLAM. 
Compared to Aachen, where the meshes were created from the images, %
the alignment between geometry and image data is imperfect. 

We follow%
~\cite{Brachmann2021ICCV}, using the top-20 images retrieved using DenseVLAD~\cite{Torii-CVPR15} descriptors extracted from the original database images and the original pseudo ground-truth provided by the 12 Scenes dataset. 
The simple MeshLoc variant with SuperGlue, applied on real images, is able to localize 94.0\% of all query images within 5cm and 5$^\circ$ 
threshold
on average over all 12 scenes. 
This is comparable to state-of-the-art methods such as Active Search~\cite{Sattler2017PAMI}, DSAC*~\cite{brachmann2020ARXIV}, and 
DenseVLAD retrieval with R2D2~\cite{revaud2019r2d2} features, which on average localize more than 99.0\% of all queries within 5cm and 5$^\circ$. 
The drop is caused by a visible misalignment between the geometry and RGB images in some scenes, \eg, apt2/living
(see Fig.~\ref{fig:12scenes_alignment_supp}), %
resulting in non-compensable errors in the 3D point positions.
Using renderings of colored meshes respectively the tricolor scheme reduces the average percentages of localized images to 65.8\% respectively 14.1\%. 
Again, the color and geometry misalignment seems the main reason for the drop when rendering colored meshes, while we did not observed such a large gap for Aachen dataset (which has 3D meshes that better align with the images). 
Still, 99.6\% / 92.7\% / 36.0\% of the images can be localized when using real images / colored renderings / tricolor renderings for a threshold of 7cm and 7$^\circ$. 
These numbers further increase to 100\% / 99.1\% / 54.2\% for 10cm and 10$^\circ$. 
Overall, our results show that using dense 3D models leads to promising results and that these representations are a meaningful alternative to SfM point clouds. 
Please see 
Sec.~\ref{sec:12scenes} %
for more 12 Scenes results.

\section{Conclusion}
In this paper, we explored dense 3D model as an alternative scene representation to the SfM point clouds widely used by feature-based localization algorithms. 
We have discussed how to adapt existing hierarchical localization pipelines to dense 3D models. 
Extensive experiments show that a very simple version of the resulting MeshLoc pipeline is able to achieve state-of-the-art results. 
Compared to SfM-based representations, using a dense scene model does not require an extensive matching step between database images when switching to a new type of local features. 
Thus, MeshLoc allows researchers to more easily prototype new types of features. 
We have further shown that promising results can be obtained when using synthetic views rendered from the dense models rather than the original images, even without adapting the used features. %
This opens up new and interesting directions of future work, \eg, more compact scene representations that still preserve geometric details, and training features for the challenging tasks of matching real images against raw scene geometry. 
The meshes obtained via classical approaches and classical, \ie, non-neural, rendering techniques that are used in this paper thereby create strong baselines for learning-based follow-up work. 
The rendering approach also allows to use techniques such as database expansion and pose refinement, which were not included in this paper due to limited space.
We released our code, meshes, and renderings. 

\noindent\textbf{Acknowledgements. }
This work was supported by the EU Horizon 2020 project RICAIP (grant agreement No. 857306), the European Regional Development Fund under project IMPACT (No. CZ.02.1.01/0.0/0.0/15\_003/0000468), a Meta Reality Labs research award under project call 'Benchmarking City-Scale 3D Map Making with Mapillary Metropolis', the Grant Agency of the Czech Technical University in Prague (No. SGS21/119/OHK3/2T/13), the OP VVV funded project CZ.02.1.01/0.0/0.0/16\_019/0000765 “Research Center for Informatics”, and the ERC-CZ grant MSMT LL1901.

\section*{Supplementary material}
\label{sec:supp_mat}
\appendix
This supplementary provides more details on the experiments presented in the Sec.~\ref{sec:experiments} of the main paper. 
In particular, Sec.~\ref{sec:implementation} provides more implementation details.
Source code for our approach and instructions on how to run it are available\footnote{\url{https://github.com/tsattler/meshloc\_release}}. 
Sec.~\ref{sec:qualitative} shows renderings of the four Aachen models (AC13-C, AC13, AC14, AC15) used in the main paper and renderings of the 12 Scenes models provided by~\cite{Valentin20163DV}. 
Sec.~\ref{sec:qualitative} furthermore provides an extended version of Tab.~\ref{tab:mesh_list} that includes details and rendering times for the 12 Scenes models. 
Sec.~\ref{sec:aachen} provides a more detailed ablation study that includes results for different inlier thresholds for RANSAC and results for R2D2~\cite{revaud2019r2d2} features and CAPS~\cite{Wang2020ECCV} descriptors. 
Sec.~\ref{sec:12scenes} shows more detailed results for the 12 Scenes dataset. 
Sec.~\ref{sec:storage_and_time} discusses the storage requirements and run time overhead of our method compared to approaches based on Structure-from-Motion .
Sec.~\ref{sec:nerf} contains discussion on the usage of neural scene representations with the MeshLoc pipeline.

We release the dense models for the Aachen Day-Night v1.1~\cite{Sattler2012BMVC,Sattler2018CVPR,Zhang2020IJCV} dataset as well as our renderings for the Aachen and the 12 Scenes~\cite{Valentin20163DV} datasets\footnote{\url{https://data.ciirc.cvut.cz/public/projects/2022MeshLoc/}}.

\section{Implementation Details}
\label{sec:implementation}
\noindent \textbf{Feature extraction and matching.} We use the image-matching-toolbox\footnote{\url{https://github.com/GrumpyZhou/image-matching-toolbox}} for both feature extraction and matching. 

\noindent \textbf{Pose estimation.} As mentioned in the main paper, we use the LO-RANSAC~\cite{Chum2002BMVC,Lebeda2012BMVC} implementation from the PoseLib~\cite{PoseLib} library with a robust Cauchy loss for non-linear refinement. 
We run RANSAC for at leat 10k and at most 100k iterations. 
For position averaging (Sec.~\ref{sec:method}), we use a volume side length of 2m and a step size of 0.25m for Aachen ($d_\text{vol}=1$, $d_\text{step}=0.25$) and a volume side length of 0.5m and a step size of 0.05m for 12 Scenes ($d_\text{vol}=0.25$, $d_\text{step}=0.05$). 
The step sizes were chosen based on the finest position thresholds used for evaluation on both datasets (0.25m respectively 0.05m). 
We did not tune any of the parameters involved in the position averaging. 

\noindent \textbf{Tricolor rendering scheme.} The tricolor lighting setup for uncolored models was inspired by the mesh visualization in the RealityCapture\footnote{\url{https://www.capturingreality.com/}} software. 
It consists of three directional lights moving with the camera frame. 
One light is slightly blue and points into the direction of the camera's vertical axis. 
The other two lights are slightly yellow, with orientations parallel with the camera's horizontal plane, at $112^{\circ}$ and $-129^{\circ}$ from the positive side of the optical axis. 
We did not tune the rendering style, but believe that this could be an interesting direction for future work. 

\noindent \textbf{Details on the query and database images.} For the Aachen Day-Night v1.1~\cite{Sattler2012BMVC,Sattler2018CVPR,Zhang2020IJCV} dataset, we use undistorted database images (where Colmap~\cite{Schoenberger2016CVPR,Schoenberger2016ECCV} was used to generate the undistorted images based on the calibration provided by the dataset). 
We did not undistort the query images but rather remove the distortion from the 2D match positions (where features were extracted from the distorted images) before pose estimatiom.
To the best of our knowledge, the query and database images of the 12 Scenes~\cite{Valentin20163DV} dataset are already undistorted. 

\section{Renderings}
\label{sec:qualitative}
Tab.~\ref{tab:mesh_list_12_scenes} is an extended version of Tab.~\ref{tab:mesh_list}. 
Besides details on the model sizes and rendering times for the Aachen Day-Night v1.1~\cite{Sattler2012BMVC,Sattler2018CVPR,Zhang2020IJCV} dataset (already presented in the main paper), we provide the same information for the 12 Scenes~\cite{Valentin20163DV} dataset. 
Note that we only evaluated the 12 Scenes dataset in its original image resolution (1296$\times$968 pixels). 

\begin{table*}[t!]
    \begin{center}
    \begin{minipage}{\textwidth}
    \caption{Statistics for the 3D meshes used for experimental evaluation as well as rendering times for different rendering styles and resolutions}
    \label{tab:mesh_list_12_scenes}
    \scriptsize{
    \begin{tabular*}{\textwidth}{@{\extracolsep{\fill}}cllrrrrr@{\extracolsep{\fill}}}
        \hline\noalign{\smallskip}
        
        & & & & & & \multicolumn{2}{c}{Render time $[\mu s]$} \\ 
        \cline{7-8}\noalign{\smallskip}
        \multicolumn{2}{l}{Model} & Style & Size [MB] & Vertices & Triangles & 800 px & full res. \\
        \noalign{\smallskip}
        \hline
        \noalign{\smallskip}
        \parbox[t]{1mm}{\multirow{9}{*}{\rotatebox[origin=c]{-90}{Aachen v1.1}}} & AC13-C & textured & 645 & $1.4 \cdot 10^{6}$ & $2.4 \cdot 10^{6}$ & 1143 & 1187 \\
        & AC13-C & tricolor & 47 & $1.4 \cdot 10^{6}$ & $2.4 \cdot 10^{6}$ & 115 & 219 \\
        & AC13 & colored & 617 & $14.8 \cdot 10^{6}$ & $29.3 \cdot 10^{6}$ & 92 & 140 \\
        & AC13 & tricolor & 558 & $14.8 \cdot 10^{6}$ & $29.3 \cdot 10^{6}$ & 97 & 152 \\
        & AC14 & colored & 1234 & $29.4 \cdot 10^{6}$ & $58.7 \cdot 10^{6}$ & 100 & 139 \\
        & AC14 & tricolor & 1116 & $29.4 \cdot 10^{6}$ & $58.7 \cdot 10^{6}$ & 93 & 205 \\
        & AC15 & colored & 2805 & $66.8 \cdot 10^{6}$ & $133.5 \cdot 10^{6}$ & 98 & 137 \\
        & AC15 & tricolor & 2538 & $66.8 \cdot 10^{6}$ & $133.5 \cdot 10^{6}$ & 97 & 160 \\
        \noalign{\smallskip}
        \hline
        \noalign{\smallskip}
        \parbox[t]{1mm}{\multirow{30}{*}{\rotatebox[origin=c]{-90}{12 Scenes}}} & apt1/kitchen & colored & 58 & $1.4 \cdot 10^{6}$ & $2.7 \cdot 10^{6}$ & - & 133 \\
        & apt1/kitchen & tricolor & 52 & $1.4 \cdot 10^{6}$ & $2.7 \cdot 10^{6}$ & - & 106 \\
        & apt1/living & colored & 99 & $2.4 \cdot 10^{6}$ & $4.7 \cdot 10^{6}$ & - & 146 \\
        & apt1/living & tricolor & 90 & $2.4 \cdot 10^{6}$ & $4.7 \cdot 10^{6}$ & - & 107 \\
        & apt2/bed & colored & 83 & $2.0 \cdot 10^{6}$ & $3.9 \cdot 10^{6}$ & - & 154 \\
        & apt2/bed & tricolor & 75 & $2.0 \cdot 10^{6}$ & $3.9 \cdot 10^{6}$ & - & 225 \\
        & apt2/kitchen & colored & 70 & $1.7 \cdot 10^{6}$ & $3.3 \cdot 10^{6}$ & - & 107 \\
        & apt2/kitchen & tricolor & 63 & $1.7 \cdot 10^{6}$ & $3.3 \cdot 10^{6}$ & - & 135 \\
        & apt2/living & colored & 136 & $3.3 \cdot 10^{6}$ & $6.4 \cdot 10^{6}$ & - & 134 \\
        & apt2/living & tricolor & 123 & $3.3 \cdot 10^{6}$ & $6.4 \cdot 10^{6}$ & - & 137 \\
        & apt2/luke & colored & 144 & $3.5 \cdot 10^{6}$ & $6.8 \cdot 10^{6}$ & - & 128 \\
        & apt2/luke & tricolor & 130 & $3.5 \cdot 10^{6}$ & $6.8 \cdot 10^{6}$ & - & 132 \\
        & office1/gates362 & colored & 122 & $3.0 \cdot 10^{6}$ & $5.7 \cdot 10^{6}$ & - & 127 \\
        & office1/gates362 & tricolor & 110 & $3.0 \cdot 10^{6}$ & $5.7 \cdot 10^{6}$ & - & 131 \\
        & office1/gates381 & colored & 171 & $4.1 \cdot 10^{6}$ & $8.1 \cdot 10^{6}$ & - & 110 \\
        & office1/gates381 & tricolor & 155 & $4.1 \cdot 10^{6}$ & $8.1 \cdot 10^{6}$ & - & 103 \\
        & office1/lounge & colored & 139 & $3.4 \cdot 10^{6}$ & $6.6 \cdot 10^{6}$ & - & 124 \\
        & office1/lounge & tricolor & 126 & $3.4 \cdot 10^{6}$ & $6.6 \cdot 10^{6}$ & - & 133 \\
        & office1/manolis & colored & 157 & $3.8 \cdot 10^{6}$ & $7.4 \cdot 10^{6}$ & - & 116 \\
        & office1/manolis & tricolor & 142 & $3.8 \cdot 10^{6}$ & $7.4 \cdot 10^{6}$ & - & 119 \\
        & office2/5a & colored & 122 & $2.9 \cdot 10^{6}$ & $5.8 \cdot 10^{6}$ & - & 117 \\
        & office2/5a & tricolor & 110 & $2.9 \cdot 10^{6}$ & $5.8 \cdot 10^{6}$ & - & 106 \\
        & office2/5b & colored & 170 & $4.1 \cdot 10^{6}$ & $8.0 \cdot 10^{6}$ & - & 108 \\
        & office2/5b & tricolor & 153 & $4.1 \cdot 10^{6}$ & $8.0 \cdot 10^{6}$ & - & 104 \\
        \noalign{\smallskip}
        \hline
    \end{tabular*}
    }
    \end{minipage}
    \end{center}
\end{table*}

\begin{figure}[t]
    \centering
    \includegraphics[width=0.95\textwidth]{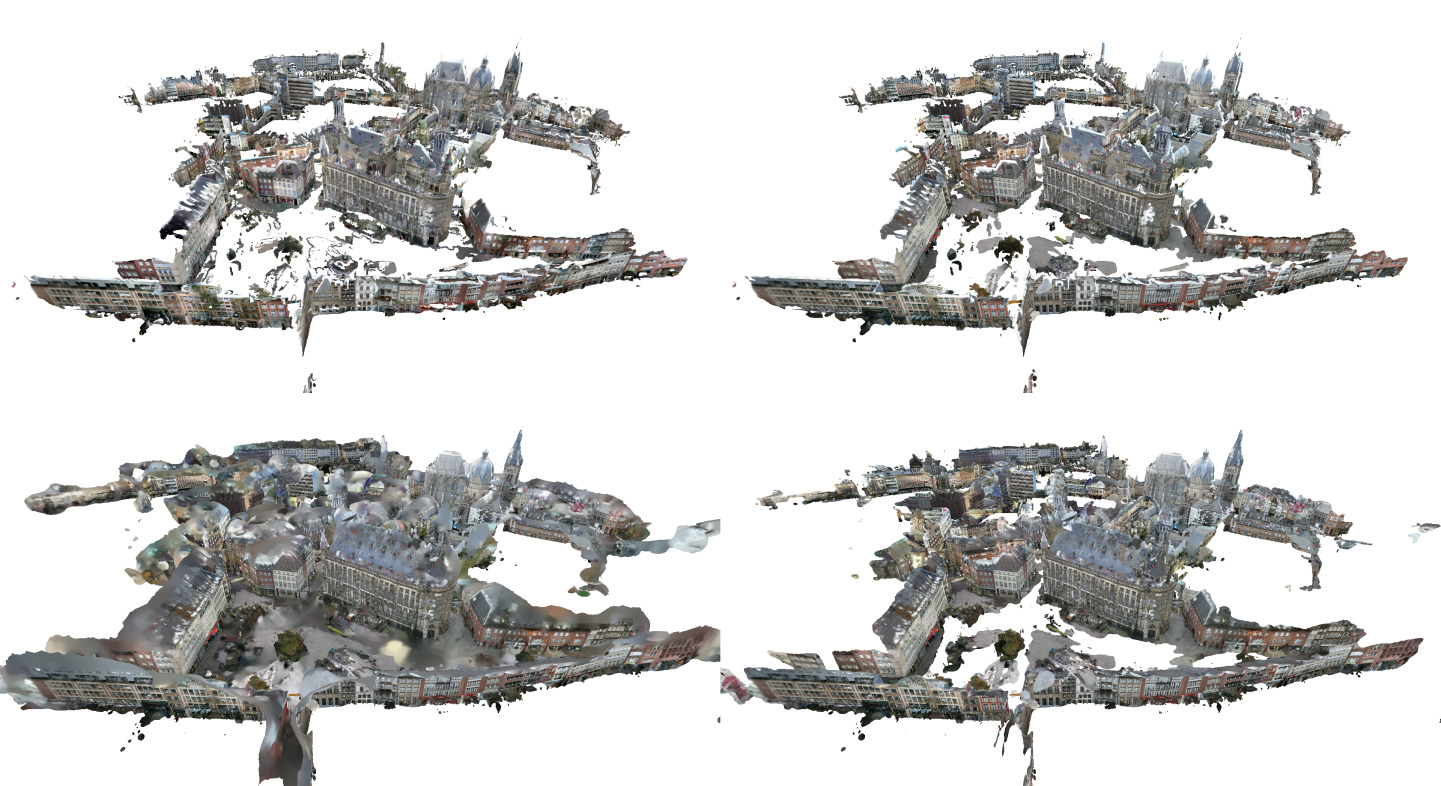}
    \caption{Comparison of the used meshes generated from Aachen database images - bird's-eye view from market square. Top-left: AC13-C (textured), top-right: AC13 (colored per vertex), bottom-left: AC14 (colored per vertex), bottom-right: AC15 (colored per vertex)}
    \label{fig:mesh_view09}
\end{figure}

\begin{figure}[t]
    \centering
    \includegraphics[width=0.95\textwidth]{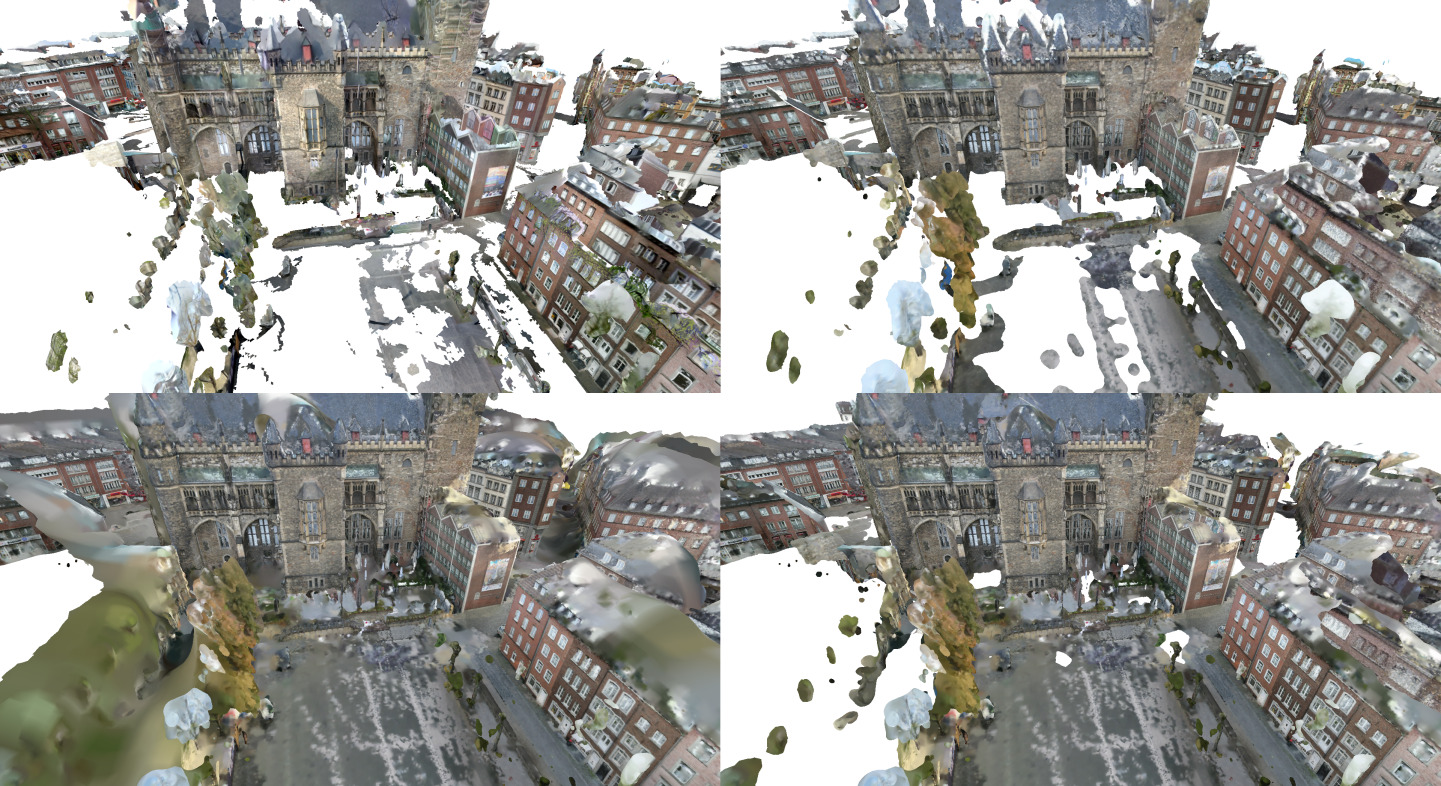}
    \caption{Comparison of the used meshes generated from Aachen database images - view at the townhall. Top-left: AC13-C (textured), top-right: AC13 (colored per vertex), bottom-left: AC14 (colored per vertex), bottom-right: AC15 (colored per vertex)}
    \label{fig:mesh_view00}
\end{figure}

\begin{figure}[t]
    \centering
    \includegraphics[width=0.95\textwidth]{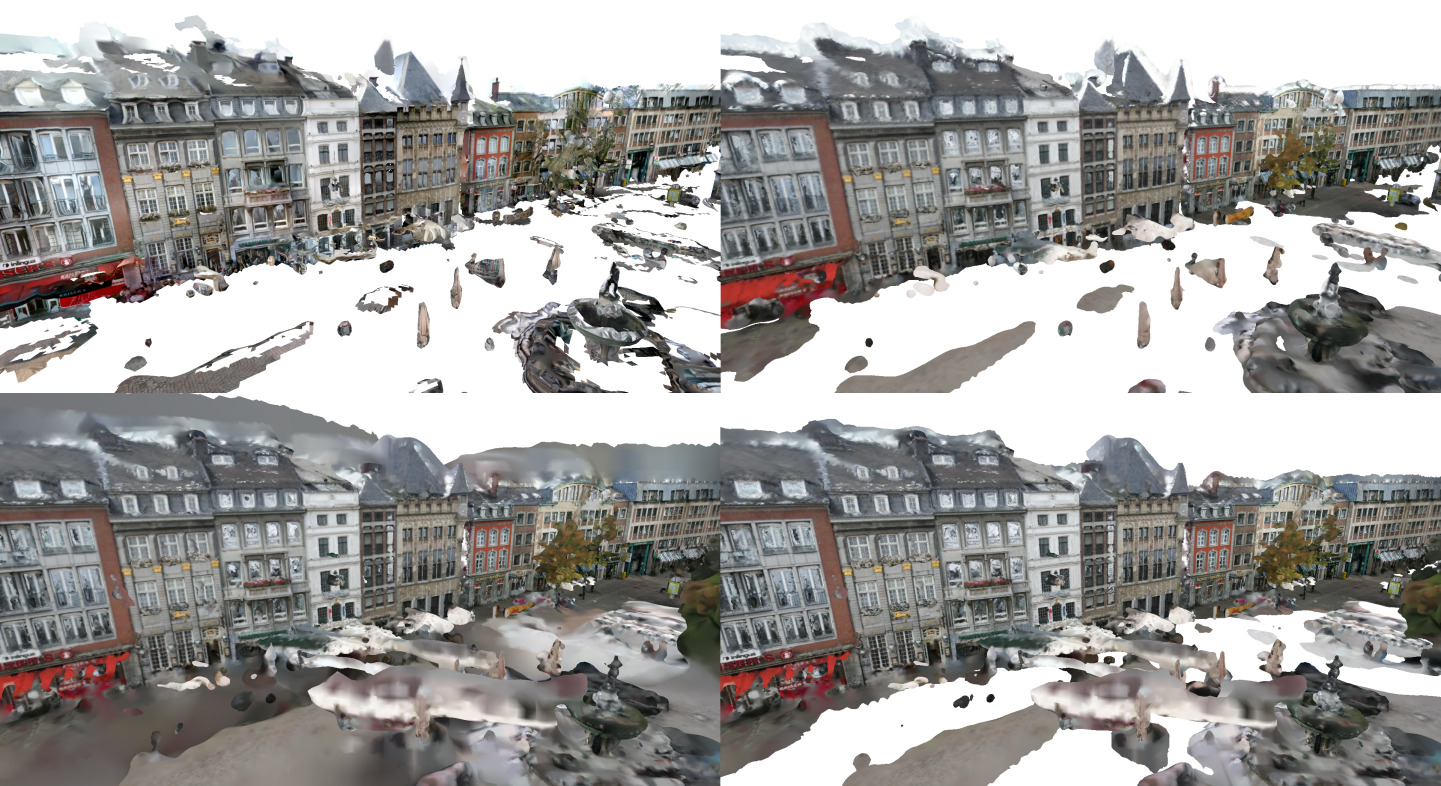}
    \caption{Comparison of the used meshes generated from Aachen database images - view at houses at market square. Top-left: AC13-C (textured), top-right: AC13 (colored per vertex), bottom-left: AC14 (colored per vertex), bottom-right: AC15 (colored per vertex)}
    \label{fig:mesh_view02}
\end{figure}

\begin{figure}[t]
    \centering
    \includegraphics[width=0.95\textwidth]{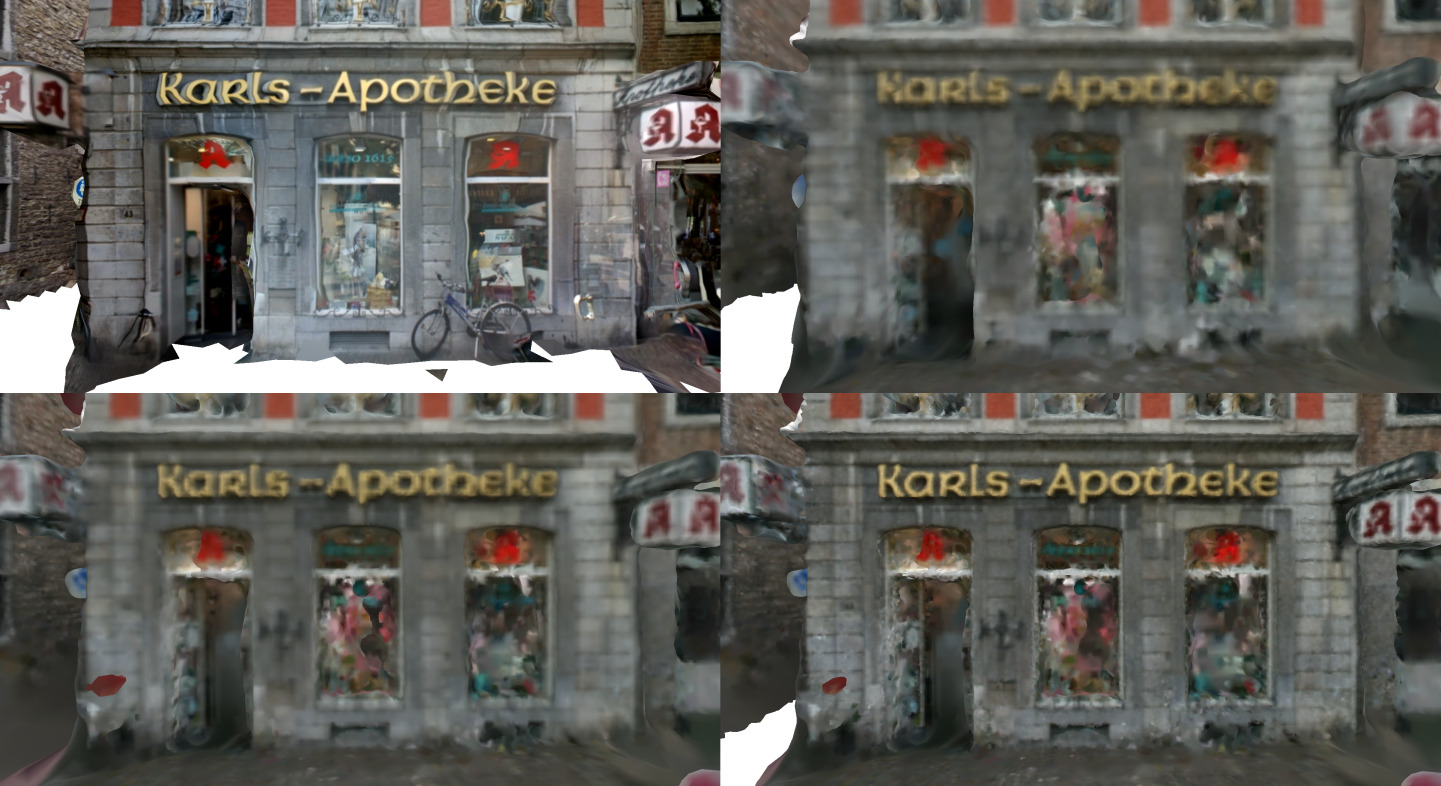}
    \caption{Comparison of the used meshes generated from Aachen database images - detail view of one of the houses. Top-left: AC13-C (textured), top-right: AC13 (colored per vertex), bottom-left: AC14 (colored per vertex), bottom-right: AC15 (colored per vertex)}
    \label{fig:mesh_view10}
\end{figure}

Figs.~\ref{fig:mesh_view09},~\ref{fig:mesh_view00},~\ref{fig:mesh_view02}, and~\ref{fig:mesh_view10} show the four Aachen models (AC13-C, AC13, AC14, AC15) from various views. 
The AC14 and AC15 models lead to sharper renderings compared to the AC13 model (the AC13-C model uses textures on top of a version of the AC13 model simplified using~\cite{jakob2015instant}). 
However, there is also more noise, in the form of floating blobs of geometry (\cf Fig.~\ref{fig:mesh_view02}), especially for the AC14 model. 
These artifacts can reduce localization performance, as noted in the main paper. 
We also attempted to create textured versions of the AC13, AC14, and AC15 but ran out of memory when applying~\cite{waechter2014let} on these significantly larger models. 


\begin{figure}[t]
    \centering
    \includegraphics[width=0.85\textwidth]{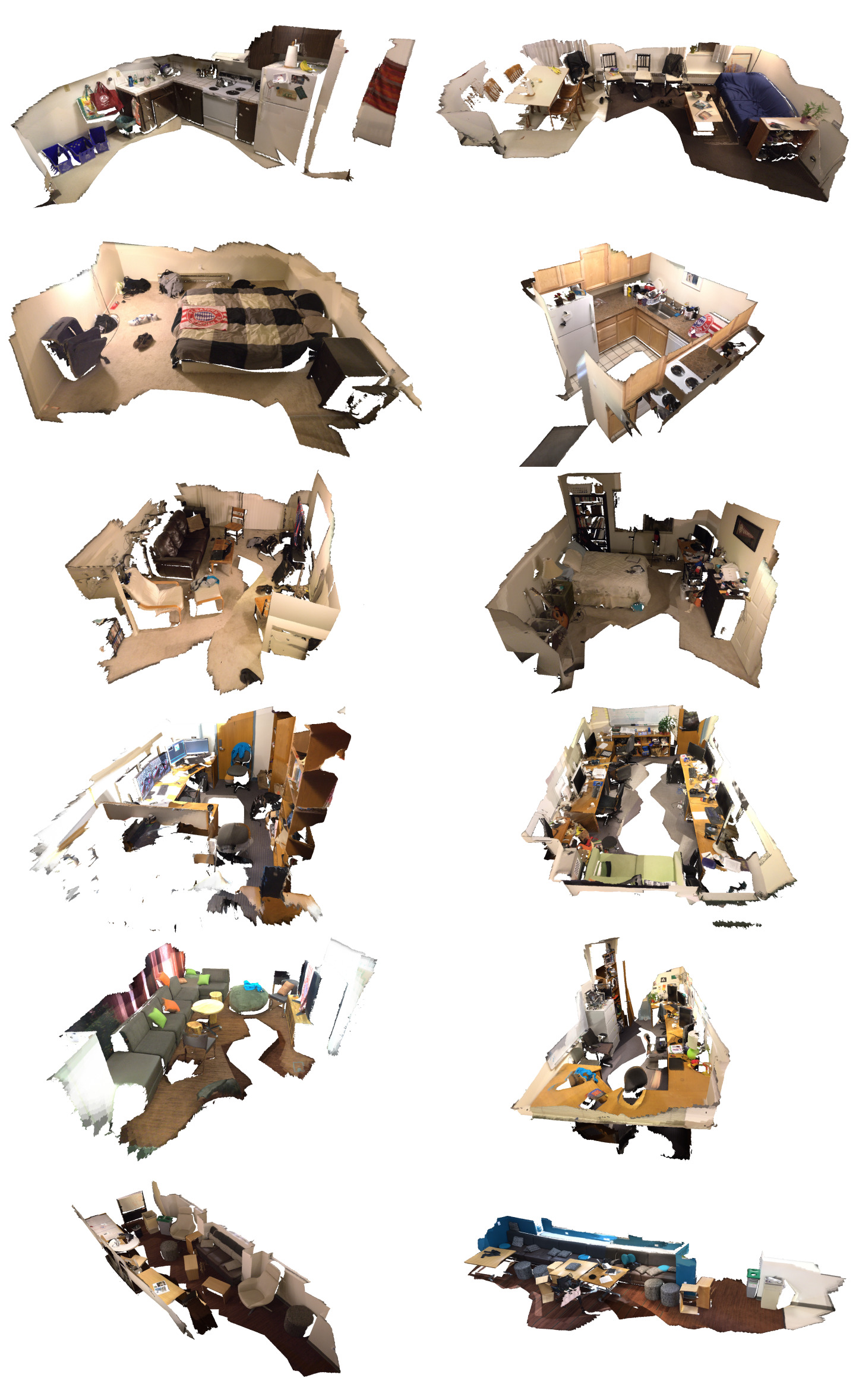}
    \caption{Comparison of the used meshes generated from 12 Scenes dataset database images. From top-left: apt1/kitchen, apt1/living (on right), apt2/bed, apt2/kitchen, apt2/living, apt2/luke, office1/gates362, office1/gates381, office1/lounge, office1/manolis, office2/5a, office2/5b}
    \label{fig:mesh_view02_12scenes}
\end{figure}

\begin{figure}[t]
    \centering
    \includegraphics[width=1.0\textwidth]{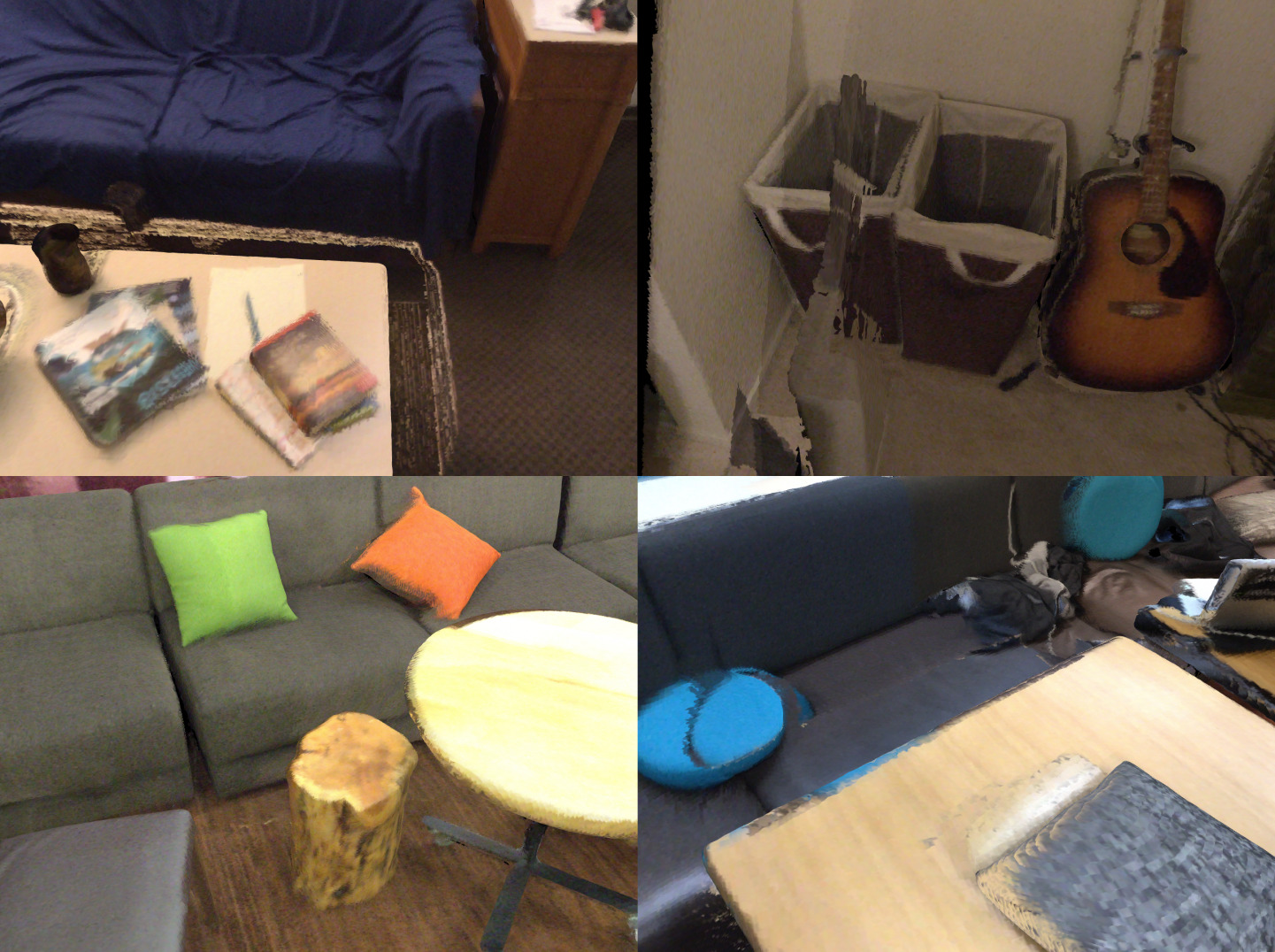}
    \caption{For some scenes of the 12 Scenes~\cite{Valentin20163DV} dataset, there are (slight) misalignments between the RGB images and the scene geometry. We show these misalignments for colored renderings from the apt1/living, apt2/luke, office1/lounge, and office2/5b scenes.}
    \label{fig:12scenes_alignment_supp}
\end{figure}

Fig.~\ref{fig:mesh_view02_12scenes} shows visualizations of the colored meshes for each of the scenes in the 12 Scenes dataset.  

\section{Experiments on Aachen Day-Night}
\label{sec:aachen}
This section provides a more detailed version of the results presented in Sec.~\ref{sec:experiments} of the main paper. %
Note that we only provide results for the case where the images have a maximum side length of 800 pixels. 
While Tab.~\ref{tab:ablation_study_real_images_800} in the paper also evaluates different features on full resolution images, the purpose of that experiment was to show that the MeshLoc pipeline can achieve a similar accuracy as SfM-based methods. 
Still, we do not consider experiments on full resolution images essential to the ablation study details presented in the following. 
Note that due to RANSAC's random nature and the fact that we re-ran the experiments for this more detailed ablation study, the results can differ (slightly) from those reported in the main paper.

Tab.~\ref{tab:aachen_real_800_supp} extends the results for reduced resolution images from Tab.~\ref{tab:ablation_study_real_images_800} in the main paper by providing results for different inlier thresholds for RANSAC and results for R2D2~\cite{revaud2019r2d2} features and CAPS~\cite{Wang2020ECCV} descriptors extracted around SuperPoint~\cite{DeTone2018CVPRWorkshops} (SP) features (denoted as CAPS+SP). 
We note that both R2D2 and CAPS+SP perform similarly well or slightly worse than the features evaluated in the main paper (which was the reason why their results were not shown in the main paper). 
In the main paper, we used the following inlier thresholds: 6 pixels for SuperGlue and LoFTR, 12 pixels for Patch2Pix+SuperGlue, and 20 pixels for Patch2Pix. 
As can be seen from Tab.~\ref{tab:aachen_real_800_supp}, the choice of the threshold is not too critical for most features. 

\begin{table}[t!]
    \begin{center}
    \begin{minipage}{\textwidth}
    \caption{Ablation study on the Aachen Day-Night v1.1 dataset~\cite{Sattler2012BMVC,Sattler2018CVPR,Zhang2020IJCV} using real images at reduced (max. side length 800 px) resolution, and depth maps rendered using the AC13 model. We evaluate different strategies for obtaining 2D-3D matches (using all individual matches (I), merging matches (M), or triangulation (T)), with and without covisibility filtering (C), and with and without position averaging (PA) for various local features. We vary the inlier threshold $t$ used in RANSAC (in pixels). We report the percentage of nighttime query images localized within 0.25m and 2$^\circ$ / 0.5m and 5$^\circ$ / 5m and 10$^\circ$ of the ground truth pose}
    \label{tab:aachen_real_800_supp}
    \tiny{
    \setlength\tabcolsep{2pt}
    \begin{tabular*}{\textwidth}{@{\extracolsep{\fill}}cccccccccc@{\extracolsep{\fill}}}
        \hline\noalign{\smallskip}
        
         & 2D- & & & SuperGlue & LoFTR & Patch2Pix & P2P & R2D2 & CAP+SP\\ 
        $t$ & \ 3D & C & PA &  (SG)~\cite{Sarlin2020CVPR} & \cite{Sun2021CVPR} &   (P2P)~\cite{Zhou2021CVPR} &  + SG~\cite{Zhou2021CVPR} & \cite{revaud2019r2d2} & \cite{Wang2020ECCV}\\  %
        \noalign{\smallskip}
\hline
\noalign{\smallskip}

\multirow{6}{*}{6.0} & I & & &
72.8/92.1/99.0 & 77.5/92.1/99.5 & 68.6/88.5/96.3 & 74.9/91.6/100.0 & 69.1/81.7/92.1 & 71.7/90.1/96.3 \\
 & I & & \checkmark & 
73.8/92.7/99.0 & 78.5/92.7/99.5 & 70.2/89.5/96.3 & 74.3/91.6/100.0 & 70.7/82.2/92.1 & 69.6/90.1/96.3  \\
 & M & & \checkmark & 
72.3/92.7/99.5 & 77.0/92.7/99.0 & 70.2/87.4/96.3 & 73.3/92.7/100.0 & 67.5/83.8/91.6 & 68.1/88.0/96.3  \\
 & M & \checkmark & \checkmark & 
71.7/92.7/99.5 & 75.9/91.6/99.5 & 69.6/88.0/97.9 & 72.8/91.6/99.5 & 67.0/88.5/99.0 & 70.7/90.1/97.9  \\
 & T & & \checkmark & 
70.7/89.0/97.4 & 74.3/90.6/98.4 & 62.3/81.2/95.3 & 73.3/89.5/97.9 & 63.9/80.6/94.8 & 62.8/85.3/95.8  \\
 & T & \checkmark & \checkmark & 
70.2/89.0/98.4 & 74.3/91.1/99.0 & 63.4/81.7/95.8 & 73.3/89.5/97.9 & 63.4/80.6/95.3 & 63.4/85.3/96.9  \\
\noalign{\smallskip}
\hline
\noalign{\smallskip}
 
\multirow{6}{*}{12.0} & I & & &
72.3/92.7/99.0 & 77.5/92.1/99.5 & 72.3/89.0/96.3 & 72.3/91.6/100.0 & 70.7/84.3/92.1 & 69.6/89.0/97.4 \\
 & I & & \checkmark & 
72.8/93.2/99.0 & 77.5/92.7/99.5 & 74.9/90.1/96.3 & 73.3/92.1/100.0 & 71.7/84.3/92.1 & 69.6/89.5/97.4  \\
 & M & & \checkmark & 
73.3/92.1/99.5 & 77.0/92.1/99.5 & 69.1/88.0/96.3 & 73.8/92.7/99.5 & 68.6/85.3/93.7 & 66.0/88.5/96.9  \\
 & M & \checkmark & \checkmark & 
72.8/92.7/99.5 & 76.4/91.6/99.5 & 69.6/90.1/97.9 & 72.8/93.2/100.0 & 68.6/85.9/99.0 & 67.0/89.0/97.4  \\
 & T & & \checkmark & 
70.2/89.5/97.9 & 75.4/92.1/98.4 & 63.4/82.7/94.8 & 70.7/90.6/97.4 & 62.8/80.1/92.7 & 62.8/85.9/95.3  \\
 & T & \checkmark & \checkmark & 
69.6/89.5/99.5 & 74.9/92.1/98.4 & 62.3/82.7/95.8 & 71.2/90.6/97.4 & 62.3/80.1/93.2 & 62.3/83.2/96.9  \\
\noalign{\smallskip}
\hline
\noalign{\smallskip}
 
\multirow{6}{*}{20.0} & I & & &
72.8/92.1/99.0 & 78.0/92.7/99.5 & 73.3/90.1/96.3 & 72.3/91.1/99.5 & 68.6/84.3/91.6 & 69.1/89.0/96.9 \\
 & I & & \checkmark & 
72.8/92.1/99.0 & 77.5/92.7/99.5 & 76.4/90.6/96.3 & 72.3/90.6/99.5 & 69.1/84.3/91.6 & 69.6/88.5/96.9  \\
 & M & & \checkmark & 
74.3/91.6/99.5 & 77.5/92.1/99.5 & 68.1/89.0/95.8 & 71.2/92.7/99.5 & 67.5/84.3/92.7 & 62.8/88.5/96.9  \\
 & M & \checkmark & \checkmark & 
72.8/91.1/99.0 & 77.5/92.1/99.5 & 69.1/91.1/97.9 & 70.2/92.1/99.5 & 68.1/86.4/97.9 & 63.9/88.5/97.9  \\
 & T & & \checkmark & 
69.1/89.5/97.4 & 72.8/91.6/98.4 & 60.7/81.7/93.7 & 70.7/89.0/97.9 & 61.3/78.5/92.1 & 58.6/83.8/95.8  \\
 & T & \checkmark & \checkmark & 
68.6/89.5/97.4 & 73.8/92.7/98.4 & 61.8/81.7/94.8 & 70.2/89.0/97.4 & 59.7/79.6/92.1 & 59.2/82.7/96.9  \\
\noalign{\smallskip}
\hline
\noalign{\smallskip}
 
\multirow{6}{*}{24.0} & I & & &
73.3/92.1/99.5 & 78.5/92.7/99.5 & 72.3/91.1/96.3 & 71.7/90.6/99.5 & 69.6/83.8/92.1 & 68.6/89.0/96.3 \\
 & I & & \checkmark & 
73.3/92.7/99.5 & 77.5/92.7/99.5 & 75.4/91.1/96.3 & 72.3/91.6/99.5 & 69.6/83.8/92.1 & 68.6/88.5/96.3  \\
 & M & & \checkmark & 
72.8/91.1/99.5 & 78.0/92.1/99.5 & 69.1/89.0/97.4 & 71.2/93.2/99.5 & 66.5/85.9/93.7 & 60.7/88.0/96.3  \\
 & M & \checkmark & \checkmark & 
71.2/91.6/99.0 & 77.5/92.1/99.5 & 67.0/90.6/97.9 & 71.2/92.7/99.5 & 66.5/85.9/96.9 & 62.8/88.5/97.9  \\
 & T & & \checkmark & 
68.1/87.4/97.4 & 73.8/91.1/98.4 & 58.6/81.7/92.7 & 70.2/90.1/97.4 & 62.8/78.0/90.6 & 57.1/82.7/96.3  \\
 & T & \checkmark & \checkmark & 
67.5/89.0/97.4 & 74.3/92.7/98.4 & 59.7/82.7/94.8 & 70.2/89.0/97.4 & 61.3/76.4/91.6 & 58.1/81.7/95.8  \\
\noalign{\smallskip}
\hline
\noalign{\smallskip}
 
\multirow{6}{*}{32.0} & I & & &
72.8/91.1/99.0 & 78.5/92.1/99.0 & 70.7/91.6/96.3 & 71.2/90.6/99.0 & 68.1/83.2/91.6 & 67.0/90.1/95.8 \\
 & I & & \checkmark & 
72.8/91.1/99.0 & 77.5/92.1/99.0 & 73.8/92.1/96.3 & 72.8/91.1/99.0 & 68.1/83.2/91.6 & 67.5/90.1/95.8  \\
 & M & & \checkmark & 
71.2/89.5/99.5 & 77.0/91.6/99.5 & 68.1/89.0/96.9 & 71.2/92.1/99.5 & 66.0/83.8/93.7 & 60.7/89.0/96.9  \\
 & M & \checkmark & \checkmark & 
70.7/90.6/99.5 & 77.0/91.6/99.5 & 68.1/90.1/97.9 & 70.2/91.1/99.5 & 66.0/83.2/95.8 & 59.7/89.0/97.4  \\
 & T & & \checkmark & 
70.2/85.9/97.4 & 72.8/89.0/98.4 & 57.6/80.6/93.2 & 69.1/89.0/96.9 & 61.3/75.9/90.1 & 56.0/81.2/95.3  \\
 & T & \checkmark & \checkmark & 
69.1/86.9/97.9 & 73.3/91.1/98.4 & 58.1/81.2/93.7 & 68.6/90.1/97.4 & 60.7/76.4/90.6 & 55.5/81.7/94.2  \\
\noalign{\smallskip}
\hline
\noalign{\smallskip}
 
\multirow{6}{*}{48.0} & I & & &
72.3/91.1/97.4 & 78.5/92.1/99.0 & 69.6/89.0/94.8 & 71.2/91.1/99.0 & 67.0/80.6/89.5 & 67.0/88.5/94.2 \\
 & I & & \checkmark & 
72.8/91.1/97.4 & 78.0/92.1/99.0 & 71.7/89.0/94.8 & 71.2/91.1/99.0 & 67.5/80.6/89.5 & 66.5/88.5/94.2  \\
 & M & & \checkmark & 
71.2/88.0/99.0 & 77.0/92.7/99.5 & 62.8/86.4/95.3 & 72.3/92.7/99.5 & 65.4/80.6/91.6 & 58.1/87.4/95.3  \\
 & M & \checkmark & \checkmark & 
70.7/89.0/99.5 & 76.4/92.7/99.0 & 64.4/89.5/96.9 & 71.7/92.7/99.5 & 63.9/83.8/94.2 & 56.5/87.4/95.3  \\
 & T & & \checkmark & 
66.5/84.3/97.4 & 72.3/89.5/97.4 & 57.1/78.5/92.1 & 65.4/86.4/97.4 & 55.5/69.1/86.4 & 52.4/76.4/91.1  \\
 & T & \checkmark & \checkmark & 
67.5/84.8/96.3 & 72.8/92.7/97.4 & 57.6/80.1/92.7 & 66.5/86.9/97.4 & 58.1/71.2/88.0 & 52.4/77.0/92.1  \\
        \noalign{\smallskip}
        \hline
    \end{tabular*}
    }
    \end{minipage}
    \end{center}
\end{table}

Tabs.~\ref{tab:ablation_study_rendered_supp1} and~\ref{tab:ablation_study_rendered_supp2} extend the results from Tab.~\ref{tab:ablation_study_rendered} by showing results obtained by the simple variant of MeshLoc, which uses all individual matches and position averaging without covisibility filtering, for rendered images. 
Besides the results for SuperGlue and Patch2Pix+SuperGlue (Patch2Pix+SG), we also show results for LoFTR, Patch2Pix, R2D2, and CAPS+ SuperPoint (CAPS+SP). 
We further show results for varying inlier thresholds.\footnote{For Tab.~\ref{tab:ablation_study_rendered}, we used the following thresholds: 12 pixels for SuperGlue on all models and for all rendering styles and 12 pixels for Patch2Pix+SuperGlue except for the colored / textured renderings, where we used a threshold of 6 pixels. } 
As can be seen from the tables, R2D2 essentially fails for the ambient occlusion and tricolor rendering styles, and also shows worse performance than the other features for colored and textured renderings. 
We further note that CAPS+SP and LoFTR typically perform worse than SuperGlue and both Patch2Pix variants for the two non-realistic rendering styles (ambient occlusion and tricolor). 
At the same time, LoFTR outperforms the other features on colored and textured renderings, reaching close to the same performance as on real images (78.5\%/93.2\%/99.5\% on real images (\cf Tab.~\ref{tab:ablation_study_real_images_800} and Tab.~\ref{tab:aachen_real_800_supp}) vs. 78.0\%/89.0\%/95.8\% for AC15 and an inlier threshold of 12 pixels (\cf Tab.~\ref{tab:ablation_study_rendered_supp2})). 
This shows that there is limited room for improvement using more realistic rendering techniques such as NeRFs~\cite{mildenhall2020nerf}. 
This result might indicate that LoFTR might focus more on textures and color patterns than on shapes and contours (only the latter two are visible in the ambient occlusion and tricolor renderings). 
The best performance for the ambient occlusion and tricolor rendering styles is typically obtained by Patch2Pix and Patch2Pix+SuperGlue. 
We attribute this good performance to both methods using a backbone network pre-trained on ImageNet~\cite{Deng2009CVPR}, which in our experience seems to lead to features that generalize quite well to unseen conditions. 
Overall, as mention in the main paper, a higher level of geometric detail (AC14 and AC15) leads to better results.

\begin{table}[t!]
    \begin{center}
    \begin{minipage}{\textwidth}
    \caption{Ablation study on the Aachen Day-Night v1.1 dataset~\cite{Sattler2012BMVC,Sattler2018CVPR,Zhang2020IJCV} using images rendered at reduced resolution (max. 800 px) from 3D meshes of different levels of detail (\cf Tab.~\ref{tab:mesh_list}) and different rendering types (textured / colored, raw geometry with ambient occlusion (AO), raw geometry with tricolor shading (tricolor)).
    We report results for a MeshLoc variant that uses individual matches and position averaging, but no covisibility filtering. We vary the inlier threshold $t$ used in RANSAC }
    \label{tab:ablation_study_rendered_supp1}
    \tiny{
    \setlength\tabcolsep{2pt}
    \begin{tabular*}{\textwidth}{@{\extracolsep{\fill}}ccccc@{\extracolsep{\fill}}}
        \hline\noalign{\smallskip}
        AC13-C: & $t$ & textured & AO & tricolor \\ 
        \noalign{\smallskip}
        \hline
        \noalign{\smallskip}
\multirow{6}{*}{SuperGlue~\cite{Sarlin2020CVPR}} 
 & 6.0 & 
71.7/91.1/99.0 & 0.5/1.0/15.7 & 6.3/19.4/39.8 \\
 & 12.0 & 
71.2/92.1/99.0 & 1.0/2.1/17.3 & 5.8/20.4/45.0 \\
 & 20.0 & 
70.2/92.7/99.0 & 1.0/2.1/17.3 & 5.2/22.0/44.5 \\
 & 24.0 & 
70.2/92.1/99.0 & 0.5/2.1/17.3 & 6.3/23.0/43.5 \\
 & 32.0 & 
70.2/92.1/99.0 & 0.5/1.6/14.7 & 7.3/19.4/39.8 \\
 & 48.0 & 
69.6/92.1/98.4 & 0.5/1.0/10.5 & 4.2/15.2/37.7 \\ \hline
\multirow{6}{*}{LoFTR~\cite{Sun2021CVPR}} 
 & 6.0 & 
74.9/90.1/99.0 & 0.0/0.0/3.7 & 1.6/11.0/37.7 \\
 & 12.0 & 
74.3/91.1/99.0 & 0.0/0.0/4.7 & 1.6/12.6/41.9 \\
 & 20.0 & 
73.8/91.1/98.4 & 0.0/0.0/6.8 & 2.6/13.6/41.9 \\
 & 24.0 & 
73.8/91.6/98.4 & 0.0/0.0/5.8 & 2.1/14.1/38.7 \\
 & 32.0 & 
74.3/91.1/98.4 & 0.0/0.0/4.2 & 1.6/13.1/36.6 \\
 & 48.0 & 
74.3/90.6/97.4 & 0.0/0.0/2.1 & 1.0/7.9/30.9 \\ \hline
\multirow{6}{*}{Patch2Pix~\cite{Zhou2021CVPR}} 
 & 6.0 & 
65.4/87.4/93.2 & 1.6/4.7/25.1 & 4.7/21.5/59.2 \\
 & 12.0 & 
66.5/87.4/94.2 & 2.1/6.8/30.4 & 9.4/28.3/63.4 \\
 & 20.0 & 
64.9/85.9/93.7 & 2.6/7.9/30.4 & 8.4/31.9/65.4 \\
 & 24.0 & 
65.4/85.3/94.2 & 4.2/10.5/30.4 & 8.4/31.9/66.5 \\
 & 32.0 & 
62.8/84.8/93.7 & 1.6/7.3/28.3 & 7.9/27.7/63.4 \\
 & 48.0 & 
62.8/83.2/93.2 & 1.0/6.3/20.4 & 6.3/22.5/52.9 \\ \hline
\multirow{6}{*}{Patch2Pix+SG~\cite{Zhou2021CVPR, Sarlin2020CVPR}} 
 & 6.0 & 
69.6/91.1/99.5 & 1.6/2.6/19.9 & 6.8/24.1/53.4 \\
 & 12.0 & 
68.1/92.1/99.5 & 1.0/2.1/22.5 & 6.8/26.2/53.9 \\
 & 20.0 & 
67.5/92.1/99.5 & 0.5/2.1/22.0 & 8.9/25.1/53.4 \\
 & 24.0 & 
67.0/91.6/99.0 & 0.5/2.6/22.0 & 7.9/25.1/52.9 \\
 & 32.0 & 
67.5/91.6/98.4 & 0.5/2.6/20.9 & 8.9/24.1/49.7 \\
 & 48.0 & 
68.1/91.6/99.0 & 0.5/1.6/19.4 & 7.9/23.0/46.1 \\ \hline
\multirow{6}{*}{R2D2~\cite{revaud2019r2d2}} 
 & 6.0 & 
58.1/73.8/83.8 & 0.0/0.0/0.0 & 0.0/0.0/0.0 \\
 & 12.0 & 
59.2/76.4/85.9 & 0.0/0.0/0.0 & 0.0/0.0/0.0 \\
 & 20.0 & 
58.6/75.9/85.9 & 0.0/0.0/0.0 & 0.0/0.0/0.5 \\
 & 24.0 & 
58.1/74.9/84.8 & 0.0/0.0/0.0 & 0.0/0.0/0.5 \\
 & 32.0 & 
56.5/73.8/83.2 & 0.0/0.0/0.0 & 0.0/0.0/1.0 \\
 & 48.0 & 
56.5/70.2/79.1 & 0.0/0.0/0.0 & 0.0/0.0/0.5 \\ \hline
\multirow{6}{*}{CAPS+SP~\cite{Wang2020ECCV,DeTone2018CVPRWorkshops}} 
 & 6.0 & 
64.9/89.0/96.9 & 1.0/7.3/40.3 & 3.7/17.8/64.4 \\
 & 12.0 & 
67.0/88.5/97.4 & 3.1/7.3/50.3 & 3.1/18.3/70.7 \\
 & 20.0 & 
66.5/89.0/96.9 & 2.6/9.4/51.3 & 3.1/18.3/70.2 \\
 & 24.0 & 
67.0/88.0/96.3 & 2.1/8.4/50.8 & 2.6/17.8/69.6 \\
 & 32.0 & 
66.5/88.0/96.9 & 3.7/9.4/46.6 & 2.6/18.8/67.5 \\
 & 48.0 & 
64.4/85.3/93.2 & 2.6/6.8/40.3 & 4.2/19.9/64.4 \\ \hline
\hline
\noalign{\smallskip}
 
 AC13: & $t$ & colored & AO & tricolor \\ \hline
\multirow{6}{*}{SuperGlue~\cite{Sarlin2020CVPR}} 
 & 6.0 & 
67.5/88.0/96.9 & 1.6/13.1/30.9 & 19.4/45.5/68.1 \\
 & 12.0 & 
70.7/90.1/97.9 & 3.1/13.6/35.1 & 23.0/49.2/69.1 \\
 & 20.0 & 
69.6/89.5/97.4 & 2.6/11.0/34.0 & 21.5/49.7/69.6 \\
 & 24.0 & 
68.6/90.1/97.9 & 3.7/12.0/35.6 & 23.0/46.6/68.1 \\
 & 32.0 & 
67.5/89.0/97.9 & 4.7/12.6/34.6 & 21.5/45.5/68.6 \\
 & 48.0 & 
66.5/88.0/96.3 & 3.7/9.9/29.8 & 18.3/35.1/61.8 \\ \hline
\multirow{6}{*}{LoFTR~\cite{Sun2021CVPR}} 
 & 6.0 & 
69.1/88.0/94.8 & 2.6/6.3/29.3 & 9.9/28.8/64.4 \\
 & 12.0 & 
72.3/86.9/94.8 & 1.6/6.8/32.5 & 15.7/36.1/63.4 \\
 & 20.0 & 
72.3/87.4/94.2 & 3.1/8.4/32.5 & 13.1/36.1/61.8 \\
 & 24.0 & 
71.7/87.4/93.7 & 3.1/8.9/29.8 & 12.6/34.6/60.2 \\
 & 32.0 & 
71.2/86.9/92.7 & 2.6/7.9/28.3 & 13.6/35.1/58.1 \\
 & 48.0 & 
68.1/84.3/89.0 & 1.6/3.1/18.8 & 8.9/27.7/49.2 \\ \hline
\multirow{6}{*}{Patch2Pix~\cite{Zhou2021CVPR}} 
 & 6.0 & 
64.4/81.7/90.1 & 5.8/20.4/49.2 & 20.9/39.8/75.4 \\
 & 12.0 & 
61.3/82.7/90.6 & 8.9/24.6/55.5 & 26.2/50.8/82.2 \\
 & 20.0 & 
60.2/81.2/92.1 & 11.0/28.3/58.6 & 30.4/58.6/80.1 \\
 & 24.0 & 
59.2/82.2/93.2 & 12.6/28.3/57.1 & 31.9/58.1/81.2 \\
 & 32.0 & 
58.6/82.7/91.6 & 13.6/27.7/57.1 & 29.8/53.4/77.5 \\
 & 48.0 & 
57.1/80.1/89.0 & 10.5/23.6/49.2 & 26.7/46.6/72.3 \\ \hline
\multirow{6}{*}{Patch2Pix+SG~\cite{Zhou2021CVPR, Sarlin2020CVPR}} 
 & 6.0 & 
71.2/92.1/97.4 & 4.7/17.8/41.9 & 22.0/47.6/72.8 \\
 & 12.0 & 
69.6/91.6/97.9 & 5.8/21.5/45.0 & 23.0/50.3/73.8 \\
 & 20.0 & 
67.5/91.6/97.9 & 5.2/23.0/45.5 & 22.0/45.5/73.8 \\
 & 24.0 & 
68.6/91.6/97.4 & 5.2/22.0/45.0 & 20.9/46.1/73.8 \\
 & 32.0 & 
69.1/91.1/96.9 & 5.2/21.5/44.5 & 20.4/43.5/72.3 \\
 & 48.0 & 
69.6/90.6/96.3 & 5.8/18.8/43.5 & 19.9/42.9/69.1 \\ \hline
\multirow{6}{*}{R2D2~\cite{revaud2019r2d2}} 
 & 6.0 & 
55.5/69.6/82.7 & 0.0/0.0/0.0 & 0.0/0.5/1.6 \\
 & 12.0 & 
57.6/71.2/83.2 & 0.0/0.0/0.0 & 1.0/1.0/2.6 \\
 & 20.0 & 
56.5/71.2/83.2 & 0.0/0.0/0.0 & 0.5/0.5/2.6 \\
 & 24.0 & 
56.0/71.2/82.7 & 0.0/0.0/0.0 & 0.5/0.5/3.1 \\
 & 32.0 & 
55.0/69.1/81.7 & 0.0/0.0/0.0 & 0.0/0.5/2.6 \\
 & 48.0 & 
52.9/66.0/75.9 & 0.0/0.0/0.0 & 0.0/0.5/2.1 \\ \hline
\multirow{6}{*}{CAPS+SP~\cite{Wang2020ECCV,DeTone2018CVPRWorkshops}} 
 & 6.0 & 
61.3/82.7/94.2 & 9.4/25.7/70.7 & 13.6/39.3/81.2 \\
 & 12.0 & 
60.7/82.7/94.8 & 9.4/30.9/75.4 & 16.8/47.1/85.9 \\
 & 20.0 & 
60.2/81.7/95.3 & 7.9/31.9/75.4 & 14.7/45.5/85.9 \\
 & 24.0 & 
58.6/81.7/95.3 & 7.9/31.9/73.8 & 14.1/45.5/83.8 \\
 & 32.0 & 
57.6/80.6/92.7 & 8.9/32.5/70.7 & 13.1/42.4/79.6 \\
 & 48.0 & 
57.1/79.1/91.1 & 7.3/27.7/64.9 & 10.5/37.2/74.3 \\ 

        \hline
    \end{tabular*}
    }
    \end{minipage}
    \end{center}
\end{table}
 
\begin{table}[t!]
    \begin{center}
    \begin{minipage}{\textwidth}
    \caption{Ablation study on the Aachen Day-Night v1.1 dataset~\cite{Sattler2012BMVC,Sattler2018CVPR,Zhang2020IJCV} using images rendered at reduced resolution (max. 800 px) from 3D meshes of different levels of detail (\cf Tab.~\ref{tab:mesh_list}) and different rendering types (textured / colored, raw geometry with ambient occlusion (AO), raw geometry with tricolor shading (tricolor)).
    We report results for a MeshLoc variant that uses individual matches and position averaging, but no covisibility filtering. We vary the inlier threshold $t$ used in RANSAC }
    \label{tab:ablation_study_rendered_supp2}
    \tiny{
    \setlength\tabcolsep{2pt}
    \begin{tabular*}{\textwidth}{@{\extracolsep{\fill}}ccccc@{\extracolsep{\fill}}}
        \hline\noalign{\smallskip}
 AC14: & $t$ & colored & AO & tricolor \\ \hline
\multirow{6}{*}{SuperGlue~\cite{Sarlin2020CVPR}} 
 & 6.0 & 
69.6/88.5/95.8 & 18.3/37.2/55.5 & 32.5/58.1/72.3 \\
 & 12.0 & 
69.1/88.5/95.8 & 21.5/40.3/58.6 & 32.5/63.4/74.9 \\
 & 20.0 & 
69.1/88.0/95.8 & 21.5/39.3/57.1 & 33.0/63.4/74.3 \\
 & 24.0 & 
69.1/88.0/95.8 & 19.4/36.1/53.4 & 32.5/62.3/74.3 \\
 & 32.0 & 
68.6/88.0/95.8 & 19.9/33.0/51.3 & 31.9/59.7/72.8 \\
 & 48.0 & 
69.6/88.0/95.8 & 20.4/31.4/46.6 & 29.3/53.4/70.2 \\ \hline
\multirow{6}{*}{LoFTR~\cite{Sun2021CVPR}} 
 & 6.0 & 
74.3/87.4/95.8 & 15.7/39.3/62.8 & 26.2/58.6/78.0 \\
 & 12.0 & 
76.4/87.4/95.3 & 19.4/42.9/66.0 & 27.7/62.3/78.5 \\
 & 20.0 & 
73.8/86.9/93.7 & 18.3/42.4/66.0 & 28.3/60.2/77.0 \\
 & 24.0 & 
73.3/85.3/93.2 & 16.2/41.4/65.4 & 29.3/60.7/75.9 \\
 & 32.0 & 
73.3/85.9/93.2 & 14.7/38.7/62.8 & 29.8/57.1/73.8 \\
 & 48.0 & 
72.8/84.3/92.1 & 13.6/35.1/57.6 & 27.2/52.9/66.5 \\ \hline
\multirow{6}{*}{Patch2Pix~\cite{Zhou2021CVPR}} 
 & 6.0 & 
62.8/81.2/92.1 & 18.8/36.1/70.2 & 30.9/58.1/80.6 \\
 & 12.0 & 
62.3/84.3/94.2 & 26.2/48.7/74.9 & 38.2/67.5/83.2 \\
 & 20.0 & 
63.9/84.3/93.2 & 26.2/51.8/74.3 & 38.2/67.5/84.3 \\
 & 24.0 & 
62.8/83.8/93.2 & 26.2/53.4/75.4 & 38.7/65.4/83.2 \\
 & 32.0 & 
63.4/85.9/92.7 & 24.1/47.6/72.8 & 37.2/66.0/83.8 \\
 & 48.0 & 
62.3/81.7/88.5 & 22.0/41.4/64.9 & 31.9/59.7/77.0 \\ \hline
\multirow{6}{*}{Patch2Pix+SG~\cite{Zhou2021CVPR, Sarlin2020CVPR}} 
 & 6.0 & 
72.3/90.6/96.9 & 19.9/40.8/63.4 & 35.6/64.9/78.5 \\
 & 12.0 & 
70.7/89.0/96.9 & 20.4/39.8/64.9 & 33.5/64.4/78.5 \\
 & 20.0 & 
70.2/89.5/96.9 & 19.4/40.3/62.8 & 32.5/61.8/78.5 \\
 & 24.0 & 
70.7/89.5/96.9 & 19.9/38.2/61.8 & 33.0/61.8/78.0 \\
 & 32.0 & 
70.2/89.0/96.9 & 19.9/37.7/59.7 & 33.5/61.8/78.0 \\
 & 48.0 & 
69.1/89.0/96.3 & 18.8/36.6/57.6 & 30.9/60.2/75.4 \\ \hline
\multirow{6}{*}{R2D2~\cite{revaud2019r2d2}} 
 & 6.0 & 
57.6/67.5/81.2 & 0.0/0.0/1.0 & 0.5/0.5/1.6 \\
 & 12.0 & 
56.5/69.6/83.8 & 0.0/0.0/1.6 & 0.5/1.0/3.1 \\
 & 20.0 & 
56.5/70.2/84.3 & 0.0/0.0/2.1 & 0.5/0.5/2.1 \\
 & 24.0 & 
55.5/69.1/82.7 & 0.0/0.0/2.6 & 1.0/1.0/2.6 \\
 & 32.0 & 
54.5/69.6/80.1 & 0.0/0.0/3.1 & 1.0/1.0/2.1 \\
 & 48.0 & 
52.9/67.0/77.5 & 0.0/0.0/2.6 & 0.0/0.5/1.0 \\ \hline
\multirow{6}{*}{CAPS+SP~\cite{Wang2020ECCV,DeTone2018CVPRWorkshops}} 
 & 6.0 & 
56.5/82.7/94.2 & 24.1/53.4/84.8 & 33.0/64.4/86.4 \\
 & 12.0 & 
60.7/84.8/95.3 & 25.1/62.8/84.8 & 30.9/65.4/90.1 \\
 & 20.0 & 
61.3/83.2/94.8 & 24.6/57.6/82.7 & 30.4/68.1/87.4 \\
 & 24.0 & 
61.3/83.2/93.7 & 24.6/55.5/82.7 & 31.4/68.6/88.0 \\
 & 32.0 & 
61.8/82.7/92.7 & 23.6/55.5/81.2 & 29.8/66.0/85.3 \\
 & 48.0 & 
59.2/79.1/89.0 & 22.5/55.5/76.4 & 25.1/62.3/80.1 \\ \hline
\hline
\noalign{\smallskip}
 
 AC15: & $t$ & colored & AO & tricolor \\ \hline
\multirow{6}{*}{SuperGlue~\cite{Sarlin2020CVPR}} 
 & 6.0 & 
71.2/89.0/96.9 & 21.5/38.2/62.3 & 37.2/55.5/70.2 \\
 & 12.0 & 
73.3/90.1/97.9 & 23.6/45.0/63.4 & 35.6/59.7/75.9 \\
 & 20.0 & 
73.3/89.5/97.4 & 24.6/40.8/59.7 & 35.6/59.7/74.9 \\
 & 24.0 & 
72.3/89.5/97.9 & 20.9/40.3/58.1 & 35.6/60.2/74.3 \\
 & 32.0 & 
71.2/88.5/96.9 & 20.4/37.2/55.5 & 35.1/54.5/70.7 \\
 & 48.0 & 
70.2/87.4/96.3 & 17.3/36.6/52.9 & 31.4/51.3/68.1 \\ \hline
\multirow{6}{*}{LoFTR~\cite{Sun2021CVPR}} 
 & 6.0 & 
75.9/88.5/95.8 & 19.9/40.8/62.8 & 32.5/53.9/74.9 \\
 & 12.0 & 
78.0/89.0/95.8 & 19.9/42.9/65.4 & 34.0/60.7/77.0 \\
 & 20.0 & 
75.9/88.0/94.2 & 19.9/41.4/61.8 & 32.5/59.7/76.4 \\
 & 24.0 & 
74.9/88.0/94.2 & 18.8/39.3/59.7 & 31.4/57.1/73.8 \\
 & 32.0 & 
74.3/87.4/94.2 & 17.3/39.3/57.1 & 30.9/52.9/69.6 \\
 & 48.0 & 
73.8/86.9/92.7 & 15.7/32.5/52.4 & 26.7/47.6/63.4 \\ \hline
\multirow{6}{*}{Patch2Pix~\cite{Zhou2021CVPR}} 
 & 6.0 & 
63.9/81.2/93.7 & 22.0/46.1/73.3 & 26.7/53.9/77.5 \\
 & 12.0 & 
63.9/84.3/93.7 & 29.8/54.5/77.5 & 35.6/60.7/80.1 \\
 & 20.0 & 
65.4/84.8/94.2 & 27.7/55.5/77.5 & 36.1/64.4/81.7 \\
 & 24.0 & 
65.4/83.8/93.7 & 26.7/53.4/76.4 & 38.2/62.8/82.2 \\
 & 32.0 & 
64.4/83.8/93.7 & 27.2/52.9/75.9 & 36.6/63.9/81.2 \\
 & 48.0 & 
63.4/81.7/89.5 & 24.6/45.5/66.0 & 35.1/57.1/74.9 \\ \hline
\multirow{6}{*}{Patch2Pix+SG~\cite{Zhou2021CVPR, Sarlin2020CVPR}} 
 & 6.0 & 
72.8/90.1/98.4 & 24.6/47.1/65.4 & 37.2/60.7/78.0 \\
 & 12.0 & 
72.3/91.6/98.4 & 24.6/49.2/67.0 & 39.3/62.3/79.6 \\
 & 20.0 & 
72.8/91.1/97.9 & 22.5/46.1/66.0 & 36.6/62.3/80.1 \\
 & 24.0 & 
71.7/91.1/97.9 & 20.4/46.1/65.4 & 35.6/61.3/81.7 \\
 & 32.0 & 
69.6/91.1/98.4 & 22.0/46.1/63.9 & 35.1/59.2/80.1 \\
 & 48.0 & 
68.1/90.6/97.9 & 21.5/43.5/63.4 & 35.1/59.7/78.0 \\ \hline
\multirow{6}{*}{R2D2~\cite{revaud2019r2d2}} 
 & 6.0 & 
56.5/72.8/82.7 & 0.0/0.0/1.0 & 0.0/0.0/1.6 \\
 & 12.0 & 
59.2/74.9/84.3 & 0.0/0.0/1.0 & 0.0/0.5/2.1 \\
 & 20.0 & 
59.2/72.8/84.3 & 0.0/0.0/1.0 & 0.0/0.5/1.6 \\
 & 24.0 & 
59.2/72.8/83.8 & 0.0/0.0/0.5 & 0.0/0.5/1.6 \\
 & 32.0 & 
57.6/70.7/83.8 & 0.0/0.0/0.5 & 0.0/0.0/1.6 \\
 & 48.0 & 
52.4/68.1/79.1 & 0.0/0.0/0.0 & 0.0/0.5/1.6 \\ \hline
\multirow{6}{*}{CAPS+SP~\cite{Wang2020ECCV,DeTone2018CVPRWorkshops}} 
 & 6.0 & 
61.8/86.9/95.3 & 28.8/58.6/83.8 & 34.0/66.0/89.0 \\
 & 12.0 & 
67.5/85.9/96.3 & 25.7/58.1/89.0 & 36.1/70.2/91.1 \\
 & 20.0 & 
67.0/85.3/95.8 & 27.7/59.7/86.4 & 34.0/69.1/89.5 \\
 & 24.0 & 
67.0/85.3/95.3 & 26.7/59.2/85.3 & 32.5/68.6/89.0 \\
 & 32.0 & 
65.4/84.8/94.2 & 27.2/57.6/84.3 & 31.4/67.0/88.5 \\
 & 48.0 & 
64.9/82.2/89.5 & 23.6/51.3/75.9 & 29.8/62.3/79.6 \\
        
        \hline
    \end{tabular*}
    }
    \end{minipage}
    \end{center}
\end{table}

\section{Experiments on 12 Scenes}
\label{sec:12scenes}
Sec.~\ref{sec:experiments} only provides results for the average number of images localized within a given error threshold for the 12 Scenes~\cite{Valentin20163DV} dataset. 
Tabs.~\ref{tab:details_12scenes_supp_5cm},~\ref{tab:details_12scenes_supp_7cm}, and~\ref{tab:details_12scenes_supp_10cm} provided detailed measurements per scene, as well as the average number of localized images, for different error thresholds. 
We use the evaluation toolkit provided by~\cite{Brachmann2021ICCV} and the original pseudo ground truth provided by the 12 Scenes dataset. 
We compare the MeshLoc results obtained with SuperGlue with the baseline methods used in~\cite{Brachmann2021ICCV}: 
Active Search~\cite{Sattler2017PAMI} and DenseVLAD+R2D2~\cite{Torii-CVPR15,HumenbergerX20Kapture,revaud2019r2d2} (DVLAD+R2D2) both use a SfM-based scene representation. 
While Active Search uses SIFT~\cite{Lowe04IJCV} features, DVLAD+R2D2 uses R2D2~\cite{revaud2019r2d2} features. 
DVLAD+R2D2 uses the same top-20 images retrieved by DenseVLAD~\cite{Torii-CVPR15} as our method. 
DVLAD+R2D2 (+D) is a variant that uses the depth images provided by the dataset to obtain 3D points instead of a SfM-based model. 
DSAC*~\cite{brachmann2020ARXIV} is a state-of-the-art scene coordinate regressor that predicts a 2D-3D match for each pixel in an image. 
We compare against an RGB-only version (DSAC*) and a version based on RGB-D query images (DSAC* (+D)).
Please see~\cite{Brachmann2021ICCV} for more details on the methods. 

We notice that on many scenes, MeshLoc performs close to the baselines when using real images instead of renderings. 
However, there are some scenes, \eg, apt1/living, apt2/luke, office1/lounge, and office2/5b, for which MeshLoc performs noticeably worse than the baselines. 
As can be seen in Fig.~\ref{fig:12scenes_alignment_supp}, we notice that for these scenes, the alignment between the RGB database images and the scene geometry is not very accurate. 
As a result, obtaining 3D points from depth maps by rendering the scene model leads to shifted 3D points, which ultimately leads to less accurate poses. 
However, as can be seen in Tab.~\ref{tab:details_12scenes_supp_10cm}, MeshLoc is still able to localize nearly all query images at reasonable precision when using either real database images or colored renderings. 

\begin{table}[t!]
    \begin{center}
    \begin{minipage}{\textwidth}
    \caption{Results on the 12 Scenes~\cite{Valentin20163DV} dataset using real images (r), images rendered from colored models (c), and images rendered using the tricolor scheme (t). 
    We report results for a MeshLoc variant that uses all individual SuperGlue matches and position averaging, but no covisibility filtering. We use an inlier threshold of 6 pixels used in RANSAC. We report the percentage of images localized within \textbf{5 cm and 5$^\circ$} of the ground truth and compare against results reported in~\cite{Brachmann2021ICCV}}
    \label{tab:details_12scenes_supp_5cm}
    \tiny{
    \setlength\tabcolsep{1pt}
    \begin{tabular*}{\textwidth}{@{\extracolsep{\fill}}l|cc|cccc|cccc|cc|c@{\extracolsep{\fill}}}
        \hline\noalign{\smallskip}
        & \multicolumn{2}{c}{apt1/} & \multicolumn{4}{c}{apt2/} & \multicolumn{4}{c}{office1/} & \multicolumn{2}{c}{office2/} & \\
     Method & kitchen & living & bed & kitchen & living & luke & gates362 & gates381 & lounge & manolis & 5a & 5b & Average \\ \hline \hline
     MeshLoc (SG) (r) & 98.0 & 
  88.2 &
 100.0 &
  97.6 &
  97.4 &
  92.9 &
  97.7 &
  93.4 &
  87.8 &
  95.5 &
  91.5 &
  88.4 & 94.0 \\ 
  MeshLoc (SG) (c) & 63.6 &
  31.0 &
  86.8 &
  92.9 &
  68.8 &
  68.8 &
  68.1 &
  78.3 &
  60.2 &
  75.7 &
  35.8 &
  60.0 &
  65.8 \\ 
  MeshLoc (SG) (t) & 17.4 &
  15.8 &
  14.7 &
  21.9 &
   5.4 &
   5.4 &
  17.6 &
  15.3 &
  12.5 &
  23.3 &
   5.6 &
  14.6 &
  14.1 \\ \hline
  Active Search & 100.0 &
    99.8 &
   100.0 &
   100.0 &
   100.0 &
    99.2 &
   100.0 &
    98.2 &
   100.0 &
   100.0 &
    98.0 &
   100.0 &
    99.6 \\ \hline
    HLoc & 100.0 &
            100.0 &
            100.0 &
            100.0 &
            100.0 &
             98.9 &
            100.0 &
             99.1 &
            100.0 &
            100.0 &
             99.6 &
            100.0 &
             99.8 \\ \hline
   DVLAD+R2D2 & 100.0 &
      100.0 &
      100.0 &
      100.0 &
      100.0 &
       99.4 &
      100.0 &
       98.4 &
      100.0 &
      100.0 &
       99.2 &
      100.0 &
       99.7 \\ \hline
    DVLAD+R2D2 (+D) & 98.0 &
 100.0 &
 100.0 &
 100.0 &
 100.0 &
  99.4 &
 100.0 &
  98.4 &
 100.0 &
  99.5 &
  96.8 &
 100.0 &
  99.3 \\ \hline
  DSAC* & 100.0 &
 100.0 &
 100.0 &
 100.0 &
 100.0 &
  97.4 &
 100.0 &
  99.3 &
 100.0 &
  97.6 &
  97.2 &
  98.8 &
  99.2 \\ \hline
  DSAC* (+D) & 100.0 &
 100.0 &
 100.0 &
 100.0 &
 100.0 &
  97.8 &
 100.0 &
  97.8 &
 100.0 &
 100.0 &
 100.0 &
 100.0 &
  99.6 \\ \hline
    \end{tabular*}
    }
    \end{minipage}
    \end{center}
\end{table}

\begin{table}[t!]
    \begin{center}
    \begin{minipage}{\textwidth}
    \caption{Results on the 12 Scenes~\cite{Valentin20163DV} dataset using real images (r), images rendered from colored models (c), and images rendered using the tricolor scheme (t). 
    We report results for a MeshLoc variant that uses all individual SuperGlue matches and position averaging, but no covisibility filtering. We use an inlier threshold of 6 pixels used in RANSAC. We report the percentage of images localized within \textbf{7 cm and 7$^\circ$} of the ground truth and compare against results reported in~\cite{Brachmann2021ICCV}}
    \label{tab:details_12scenes_supp_7cm}
    \tiny{
    \setlength\tabcolsep{1pt}
    \begin{tabular*}{\textwidth}{@{\extracolsep{\fill}}l|cc|cccc|cccc|cc|c@{\extracolsep{\fill}}}
        \hline\noalign{\smallskip}
        & \multicolumn{2}{c}{apt1/} & \multicolumn{4}{c}{apt2/} & \multicolumn{4}{c}{office1/} & \multicolumn{2}{c}{office2/} & \\
     Method & kitchen & living & bed & kitchen & living & luke & gates362 & gates381 & lounge & manolis & 5a & 5b & Average \\ \hline \hline
     MeshLoc (SG) (r) & 
 100.0 &
 100.0 &
 100.0 &
  99.0 &
 100.0 &
  99.7 &
 100.0 &
  98.7 &
 100.0 &
  99.9 &
  98.6 &
  99.8 &
  99.6 
\\
 MeshLoc (SG) (c) & 
  95.8 &
  83.0 &
 100.0 &
  98.6 &
  98.6 &
  96.2 &
  96.9 &
  97.6 &
  95.1 &
  98.0 &
  69.2 &
  83.2 &
  92.7 
\\ 
 MeshLoc (SG) (c) & 
  42.6 &
  40.4 &
  37.7 &
  56.2 &
  15.5 &
  21.8 &
  46.4 &
  36.2 &
  28.1 &
  64.6 &
  12.5 &
  29.9 &
  36.0 
\\ \hline
 Active Search & 
   100.0 &
    99.8 &
   100.0 &
   100.0 &
   100.0 &
    99.8 &
   100.0 &
    99.0 &
   100.0 &
   100.0 &
    98.6 &
   100.0 &
    99.8 
\\ \hline
 HLoc & 
            100.0 &
            100.0 &
            100.0 &
            100.0 &
            100.0 &
             99.8 &
            100.0 &
             99.9 &
            100.0 &
            100.0 &
            100.0 &
            100.0 &
            100.0 
\\ \hline
 DVLAD+R2D2 & 
      100.0 &
      100.0 &
      100.0 &
      100.0 &
      100.0 &
      100.0 &
      100.0 &
       99.1 &
      100.0 &
      100.0 &
       99.6 &
      100.0 &
       99.9 
\\ \hline
 DVLAD+R2D2 (+D) & 
 98.0 &
 100.0 &
 100.0 &
 100.0 &
 100.0 &
 100.0 &
 100.0 &
 98.7 &
 100.0 &
 99.7 &
 97.4 &
 100.0 &
 99.5 
\\ \hline
 DSAC* & 
 100.0 &
 100.0 &
 100.0 &
 100.0 &
 100.0 &
  98.1 &
 100.0 &
  99.7 &
 100.0 &
  99.1 &
 100.0 &
 100.0 &
  99.7 
\\ \hline
 DSAC*  (+D) & 
 100.0 &
 100.0 &
 100.0 &
 100.0 &
 100.0 &
  97.9 &
 100.0 &
  99.0 &
 100.0 &
 100.0 &
 100.0 &
 100.0 &
  99.7 
\\ \hline
    \end{tabular*}
    }
    \end{minipage}
    \end{center}
\end{table}

\begin{table}[t!]
    \begin{center}
    \begin{minipage}{\textwidth}
    \caption{Results on the 12 Scenes~\cite{Valentin20163DV} dataset using real images (r), images rendered from colored models (c), and images rendered using the tricolor scheme (t). 
    We report results for a MeshLoc variant that uses all individual SuperGlue matches and position averaging, but no covisibility filtering. We use an inlier threshold of 6 pixels used in RANSAC. We report the percentage of images localized within \textbf{10 cm and 10$^\circ$} of the ground truth and compare against results reported in~\cite{Brachmann2021ICCV}}
    \label{tab:details_12scenes_supp_10cm}
    \tiny{
    \setlength\tabcolsep{1pt}
    \begin{tabular*}{\textwidth}{@{\extracolsep{\fill}}l|cc|cccc|cccc|cc|c@{\extracolsep{\fill}}}
        \hline\noalign{\smallskip}
        & \multicolumn{2}{c}{apt1/} & \multicolumn{4}{c}{apt2/} & \multicolumn{4}{c}{office1/} & \multicolumn{2}{c}{office2/} & \\
     Method & kitchen & living & bed & kitchen & living & luke & gates362 & gates381 & lounge & manolis & 5a & 5b & Average \\ \hline \hline
     MeshLoc (SG) (r) & 
 100.0 &
 100.0 &
 100.0 &
 100.0 &
 100.0 &
 100.0 &
 100.0 &
  99.7 &
 100.0 &
 100.0 &
  99.8 &
 100.0 &
 100.0 
\\
 MeshLoc (SG) (c) & 
 100.0 &
  99.8 &
 100.0 &
 100.0 &
 100.0 &
  99.8 &
 100.0 &
  99.9 &
  99.7 &
 100.0 &
  92.2 &
  97.8 &
  99.1 
\\ 
 MeshLoc (SG) (t) & 
  69.2 &
  60.6 &
  49.0 &
  76.7 &
  26.6 &
  40.5 &
  75.1 &
  52.2 &
  47.1 &
  84.8 &
  20.3 &
  47.9 &
  54.2 
\\ \hline
 Active Search & 
   100.0 &
   100.0 &
   100.0 &
   100.0 &
   100.0 &
    99.8 &
   100.0 &
    99.2 &
   100.0 &
   100.0 &
    99.4 &
   100.0 &
    99.9 
\\ \hline
 HLoc & 
            100.0 &
            100.0 &
            100.0 &
            100.0 &
            100.0 &
            100.0 &
            100.0 &
            100.0 &
            100.0 &
            100.0 &
            100.0 &
            100.0 &
            100.0 
\\ \hline
 DVLAD+R2D2 & 
      100.0 &
      100.0 &
      100.0 &
      100.0 &
      100.0 &
      100.0 &
      100.0 &
       99.1 &
      100.0 &
      100.0 &
      100.0 &
      100.0 &
       99.9 
\\ \hline
 DVLAD+R2D2 (+D) & 
 98.0 &
 100.0 &
 100.0 &
 100.0 &
 100.0 &
 100.0 &
 100.0 &
 98.7 &
 100.0 &
 99.7 &
 98.4 &
 100.0 &
 99.6 
\\ \hline
 DSAC* & 
 100.0 &
 100.0 &
 100.0 &
 100.0 &
 100.0 &
  98.4 &
 100.0 &
  99.9 &
 100.0 &
  99.6 &
 100.0 &
 100.0 &
  99.8 
\\ \hline
 DSAC*  (+D) & 
 100.0 &
 100.0 &
 100.0 &
 100.0 &
 100.0 &
  98.1 &
 100.0 &
 100.0 &
 100.0 &
 100.0 &
 100.0 &
 100.0 &
  99.8 
\\ \hline
    \end{tabular*}
    }
    \end{minipage}
    \end{center}
\end{table}

\section{Run time and storage consumption}
\label{sec:storage_and_time}
SfM-based methods store 3D point positions and visibility information.
Based on numbers from the authors, this is 192MB for LoFTR and 84MB for SuperGlue for Aachen, while our tricolor AC13-C model only requires 47MB (plus an additional 0.18MB for storing camera poses). 
Storing the descriptors of the SfM points is often more expensive than storing the original images (7.36GB \textit{vs.} between 4.5 and 5GB for Aachen). 
Rendering images rather than storing them further reduces memory requirements: the colored AC15 model uses 2.8GB (Tab.~\ref{tab:mesh_list}) at a similar pose accuracy (Tab.~\ref{tab:ablation_study_rendered}). 
Using dense meshes can thus reduce memory consumption. 

For each of the top-$k$ retrieved images, MeshLoc renders a depth map (and potentially an image) in time $T_\text{R}$, extracts features from the retrieved database image in time $T_\text{db}$, and matches these features against features extracted from the query image in time $T_\text{M}$. 
Ignoring the retrieval stage (which is shared by methods based on SfM point clouds), this results in an overall time of $T_\text{q} + k \cdot (T_\text{R} + T_\text{db} + T_\text{M})$, where $T_\text{q}$ is the time needed for query feature extraction.
SfM-based methods need either $T_\text{q} + k \cdot T_\text{M}$ when using pre-extracted features or $T_\text{q} + k \cdot (T_\text{db} + T_\text{M})$ when computing features on the fly to save memory.
LoFTR and Patch2Pix(+SG) extract their features as part of the matching process as both are based on densely extracted features, 
resulting in times $k \cdot (T_\text{R} + T_\text{M})$ (MeshLoc) respectively $k \cdot T_\text{M}$ (SfM). 
For Aachen, we have $k=50$ and $T_\text{R}\approx0.2$ms. 
Using Patch2Pix+SG and AC15, MeshLoc thus requires less memory than SfM-based methods at an overhead of only 10ms, while performing similar to using the orig.  images (Tab.~\ref{tab:ablation_study_rendered}).

\section{Using neural-based rendering techniques}
\label{sec:nerf}
Our implementation uses an OpenGL rendering pipeline, which is a well-matured technology optimized for real-time performance and use of GPUs, allowing us to render on-the-fly.
The MeshLoc pipeline can readily be used with any rendering technique that provides images and depth maps, such as Neural Radience Field (NeRF).
Preliminary experiments with a recent NeRF implementation~\cite{mueller2022instant} resulted in realistic renderings for the 12 Scenes dataset~\cite{Valentin20163DV}. 
We did not obtain good depth maps using NeRF, but upcoming work, \eg,~\cite{sun2022neuconw}, promises to solve this issue. 
Another issue is the scalability of the neural scene representations, which is being targeted in several recent publications, \eg,~\cite{turki2021meganerf, xiangli2021citynerf}.
We're excited by the recent progress of Neural Rendering and we believe, that the current interest of the community will push the performance of Neural Rendering at the level of standard pipelines in terms of quality, speed and will surpass the standard SfM meshes in terms of memory efficiency in the coming years.
We're watching the state of Neural Rendering and want to pursue their use for the task in following publications.

\clearpage
\bibliographystyle{splncs04}
\bibliography{paper}
\end{document}